\documentclass[conf]{new-aiaa}
\usepackage[utf8]{inputenc}

\usepackage{graphicx}
\usepackage{amsmath}
\usepackage[version=4]{mhchem}
\usepackage{siunitx}
\usepackage{longtable,tabularx}
\usepackage{multicol}
\usepackage[dvipsnames]{xcolor}
\definecolor{orange}{HTML}{B24A17}
\hypersetup{
    colorlinks,
    linkcolor=orange,
    citecolor=orange,
    urlcolor=orange
}

\newcommand{\fakecaption}{%
  \vskip0.5\baselineskip
  \refstepcounter{table}%
}

\title{
Learning Interpretable Models of Aircraft Handling Behaviour by Reinforcement Learning from Human Feedback
\vspace{-0.45cm}
}

\author{Tom Bewley\footnote{PhD Student, Department of Engineering Mathematics. Email: tom.bewley@bristol.ac.uk.
}, Jonathan Lawry\footnote{Professor, Department of Engineering Mathematics. Email: j.lawry@bristol.ac.uk.
} and Arthur Richards\footnote{Professor, Department of Aerospace Engineering and Bristol Robotics Laboratory. Email: arthur.richards@bristol.ac.uk. 
\vspace{0.15cm}
}}
\affil{University of Bristol, Bristol, United Kingdom}

\begin{document}

\maketitle

\begin{abstract}
\vspace{-0.025cm}
We propose a method to capture the handling abilities of fast jet pilots in a software model via reinforcement learning (RL) from human preference feedback. We use pairwise preferences over simulated flight trajectories to learn an interpretable rule-based model called a reward tree, which enables the automated scoring of trajectories alongside an explanatory rationale. We train an RL agent to execute high-quality handling behaviour by using the reward tree as the objective, and thereby generate data for iterative preference collection and further refinement of both tree and agent.
Experiments with synthetic preferences show reward trees to be competitive with uninterpretable neural network reward models on quantitative and qualitative evaluations.
\vspace{-0.7cm}
\end{abstract}

\section{Introduction}
\vspace{0.2cm}

Pilots of fast jet aircraft require exceptional handling abilities, acquired over years of advanced training.
%which are typically acquired through advanced training with time-intensive feedback from expert instructors. Access to instructor time is a major bottleneck on training progression, which motivates the development of intelligent software tools to support and accelerate the process through automated scoring, feedback and correction of trainee pilot behaviour.
There would be significant practical value in a method capable of distilling the skills, knowledge and preferences of pilots and other domain experts into a software model that captures realistic handling behaviour.
The scalability of such a model would make it useful for strategic planning exercises, training,
% of a range of operational roles,
and development and testing of other software systems.
% By amplifying the effective capacity of human expertise in this way, it would be possible to derive the maximum possible benefit from this valuable but highly constrained resource.
This would enable greater return from the scarce resource of human piloting expertise.

This vision faces the practical challenge of accurately eliciting the desired knowledge for codification into an automated system.
As in many contexts requiring intuitive decision-making and rapid motor control,
% %instructor preferences
% the preferences of experts about what constitutes safe and effective aircraft handling behaviour are in large part tacit, and thus defy direct scrutiny or exact verbal description \cite{sternberg1999tacit}. Put simply:
% %instructors
experts
know good handling when they see it, but cannot always express \textit{why} in formal or linguistic terms~\cite{sternberg1999tacit}.\footnote{
% \setstretch{0.7}
This statement certainly underestimates the rich complexity of
%instructor
human
expertise; in reality, an
%instructor's
expert's
mental model is likely to be partly tacit and partly explicit. The strategy taken in this paper is to operate \textit{as if} the mental model were $100\%$ tacit, and explore what can be achieved under such a strong restriction. Real-world applications would likely benefit from combining this approach with some amount of hand-coded expert knowledge.
}
An explicit knowledge elicitation strategy would also likely be time-intensive, as would any approach relying on expert demonstration.
% and large space of plausible reward functions, which may reflect the divergent priorities and stylistic preferences of aeronautical experts.
% Hand-coding at risk of `flattening out' this potentially valuable diversity.
This motivates a learning-based approach using a sparser data source.
In light of the importance of transparency for safety-critical aviation applications \cite{rudin2019stop,brunton2021data}, it is crucial that any such approach learns an \textit{interpretable} (i.e. human-readable and comprehensible) model of the expert knowledge, to facilitate trust and verification.

This paper proposes a possible solution to this brief.
We use an artificial reinforcement learning (RL) agent to generate a dataset of simulated flight trajectories, then consult an expert to obtain pairwise preferences over those trajectories, indicating which is preferred as a solution to a given task of interest.
Pairwise preference elicitation is known to be robust and time-efficient, and provides a basis for combining data from multiple experts without the challenge of agreeing a common scoring system.
We then use statistical learning algorithms to construct an interpretable explanatory model of the gathered preferences in the form of a rule-based tree structure.
In turn, the tree is used as a reward function to train the agent to generate higher-quality trajectories, and the process is iterated to convergence.
The end result is two distinct outputs that could form valuable components of future
%pilot training
planning, training and development
software:
\begin{enumerate}
    \vspace{0.1cm}
    \item A tree-structured reward function (referred to as a \textit{reward tree}), which captures tacit expert preferences in a human-readable form. A reward tree may be used for consistent automated scoring of flight trajectories executed by
    %trainee pilots.
    human or artificial pilots in a way that is aligned with the judgement that
    %an instructor
    the original expert
    would have made, alongside an explanatory rationale that
    %the trainee can leverage to improve their future performance.
    can be used to justify, verify and improve handling behaviour.
    \vspace{0.1cm}
    \item An RL agent capable of
    %demonstrating
    executing
    high-quality handling behaviour with respect to the objective specified by the reward tree,
    %and locally correcting suboptimal trajectories executed by trainees.
    for use in simulation (e.g. as a scalable demonstrator for pilot training).
    %By adopting an agent architecture that uses explicit planning, we can trace the causal origins of the agent's actions back through the tree structure to the underlying preference dataset, providing a deep form of explanation.
    \vspace{0.1cm}
\end{enumerate}

RL for aviation is well-studied, with impressive results using neural networks (NNs),
but typically requires reward to be heuristically (thus potentially erroneously) defined rather than flexibly learnt, and lacks interpretability of the underlying model.
Our proposal is a model that is effective, interpretable and learnable from human feedback.

Having explored the fundamentals of preference-based reward tree learning in some simple benchmark cases in prior work \cite{bewley2022interpretable}, our primary aim in this paper is to perform a proof-of-concept application to a more complex problem motivated by a real industrial need.
In particular, we seek to compare the fast jet handling performance of RL agents trained using learnt reward trees to those using deep NNs as their reward learning models, which represent the state-of-the-art from prior work \citep{christiano2017deep,lee2021pebble} but lack the crucial interpretability properties required for high-stakes applications.
To perform this comparative evaluation at scale, we require a large quantity of preference data for multiple fast jet handling tasks. Since collecting such data from real pilots and aviation experts would be costly and logistically complex, our proof-of-concept experiments use synthetic preferences with respect to nominal \textit{oracle} evaluator functions of varying complexity.
Using oracles as a proxy for human evaluators is popular \citep{griffith2013policy,christiano2017deep,reddy2020learning,lindner2021information,lee2021pebble} as it enables scalable systematic comparison, with the ability to quantify performance (and in our case, appraise learnt trees) in terms of the reconstruction of a known ground truth. However, emulating a human with an oracle that responds with perfect rationality is unrealistic \citep{lee2021bpref}. For this reason, we also examine the performance impacts of noisy and myopic oracles, and a restricted data budget.

This paper describes experiments with synthetic oracle preference data for three fast jet handling tasks.
We find that reward trees can be competitive with NN reward models in both quantitative and qualitative aspects of learning performance, while having the advantage of human-readability.
Our secondary contributions include improvements to our original learning algorithm, and an illustrative analysis of learnt tree structures to demonstrate their interpretability.

\vspace{-0.4cm}
\section{Background and Related Work}
\label{sec:related_work}
\vspace{0.2cm}

\textbf{Markov Decision Processes (MDPs)}\quad In this canonical formulation of sequential decision making \cite{sutton2018reinforcement}, the state of a system at time $t$, $s_t\in\mathcal{S}$, and the action of an agent, $a_t\in\mathcal{A}$, condition the successor state $s_{t+1}$ according to dynamics $D:\mathcal{S}\times\mathcal{A}\rightarrow\Delta(\mathcal{S})$.
% ($\Delta(\cdot)$ denotes the set of all probability distributions over a set).
A reward function $R:\mathcal{S}\times\mathcal{A}\times\mathcal{S}\rightarrow\mathbb{R}$ then outputs a scalar reward $r_{t+1}$ given $s_t$, $a_t$ and $s_{t+1}$. RL algorithms use exploratory data collection to learn action-selection policies $\pi:\mathcal{S}\rightarrow\Delta(\mathcal{A})$ for MDPs, with the goal of maximising the expected discounted sum of future reward, $\mathbb{E}_{D,\pi}\sum_{h=0}^\infty\gamma^h r_{t+h+1},\gamma\in[0,1]$.

\textbf{RL for Aviation}\quad RL has seen widespread adoption in aviation \cite{azar2021drone,liu2022reinforcement,razzaghi2022survey}.
It has been used to learn landing \cite{tang2020deep} and aerobatics \cite{clarke2020deep} behaviours for fixed-wing aircraft, as an alternative or supplement to learning from costly human demonstrations \cite{morales2004learning,cao2022demonstration}.
Other work has retained a focus on learning from and with humans, using RL to predict pilot interactions in an airspace \cite{yildiz2014predicting}, implement shared autonomy for single aircraft control \cite{vemuru2019reinforcement}, and simulate student learning dynamics in pilot training \cite{van2017towards}.
Our proposal is an alternative integration of humans into the RL process.

% \vspace{0.1cm}
\textbf{Reward Learning}\quad In the usual MDP framing, reward is an immutable property of the environment, which belies the practical fact that AI objectives originate in the uncertain goals and preferences of humans \citep{russell2019human}. Reward learning~\citep{leike2018scalable} replaces hand-specified reward functions with models learnt from feedback signals
%which can be formalised as implicit choices over alternative distributions of state-action trajectories \citep{jeon2020reward}. These cues may consist of
such as demonstrations \citep{ng2000algorithms}, scalar evaluations \citep{knox2008tamer}, approval labels \citep{griffith2013policy}, and corrections \citep{bajcsy2017learning}.
In this work, we adopt the preference-based approach \cite{christiano2017deep}, in which a human observes pairs of agent trajectories and expresses which of each pair they prefer as a solution to a given task of interest. A reward function is learnt to reconstruct the pattern of preferences. This approach is popular \citep{wirth2016model,sadigh2017active,lee2021pebble,cao2021weak} and has a firm psychological basis. Experimental results indicate that humans find it easier to make relative (\textit{vs.} absolute) quality judgements \citep{kendall1975kendall,wilde2020improving} and exhibit lower variance when doing so \citep{guo2018experimental}.
% This may be because it avoids the need to maintain an absolute scale in working memory, which is liable to induce bias as it shifts over time \citep{eric2007active}.
%Accurate and computationally efficient algorithms exist for translating \citep{hullermeier2008label}.
Providing preferences also incurs less cognitive burden than providing demonstrations \cite{ibarz2018reward} and may enable more fine-grained distinctions \cite{biyik2022learning}.

% Reward modelling motivation: reward/policy separation itself brings interpretability benefit
% The default use of NNs for reward learning severely limits interpretability; reward trees provide a possible solution.

% \vspace{0.1cm}
\textbf{Explainable RL (XRL)}\quad Surveys of methods for making RL understandable to humans \citep{puiutta2020explainable,heuillet2021explainability} divide between intrinsic approaches, which imbue agents with structure such as object representations \citep{zhu2018object} or symbolic policy primitives \citep{verma2018programmatically}, and post hoc analyses of learnt policies \citep{zahavy2016graying}, including feature importance \citep{huber2019enhancing}. Spatiotemporal scope varies from the local explanation of single actions \citep{van2018contrastive} to the summary of entire policies via prototype trajectories \citep{amir2018highlights} or states \citep{huang2018establishing}. While most post hoc methods focus on fixed policies, some investigate the dynamics of agent learning \citep{dao2018deep,bewley2022summarising}.

% \vspace{0.1cm}
\textbf{Interpretable Reward Functions}\quad At the intersection of reward learning and XRL lie efforts to understand reward functions and their effects on action selection. While this area is \textit{``less developed"} than others in XRL \citep{glanois2021survey}, there exist both intrinsic approaches, which decompose rewards into semantic components \citep{juozapaitis2019explainable} or optimise for sparsity \citep{devidze2021explicable}, and post hoc approaches, which apply feature importance analysis \citep{russell2019explaining}, counterfactual probing \citep{michaud2020understanding}, or simplifying transformations \citep{jenner2022preprocessing}.
% \citet{sanneman2022empirical} use human-oriented metrics to compare the efficacy of reward explanation techniques.
% \citet{tabrez2019explanation} infers and explains missing information from human reward function
%\citet{guan2021widening} explanation-augmented feedback
Reward tree learning is an intrinsic approach, as the rule structure is inherently readable.

% \vspace{0.1cm}
\textbf{Trees in RL}\quad Trees have a long history in RL \citep{chapman1991input,dvzeroski1998relational,pyeatt2003reinforcement}.
Their use is increasingly given an XRL motivation. Applications again divide into intrinsic methods, where an agent's policy \citep{pmlr-v108-silva20a}, value function \citep{liu2018toward} or dynamics model \citep{jiang2019experience} is a tree, and post hoc tree approximations of an existing agent's policy \citep{bastani2018verifiable} or transition statistics \citep{bewley2022summarising}. Related to our focus on learning from humans, \citet{cobo2012automatic} learn tree-structured MDP abstractions from demonstrations
%\citet{lafond2013cognitive} use trees to model expert judgements for naval air defence;
and \citet{tambwekar2021specifying}
%warm-start RL by distilling
distil
a differentiable tree policy from natural language. 
While \citet{sheikh2022learning} use tree evolution to learn dense intrinsic rewards from sparse environment ones, our prior work \cite{bewley2022interpretable} is the first to learn reward trees in the absence of any ground-truth reward signal, and the first to do so from human feedback.

\vspace{-0.4cm}
\section{Aircraft Handling Environment, Tasks and Oracles}
\label{sec:env_and_tasks}
\vspace{0.2cm}

\begin{figure}[h!]
\centering
\includegraphics[width=\textwidth]{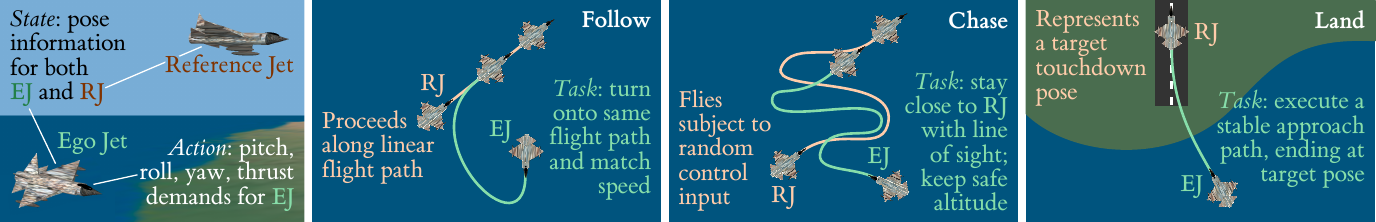}
\caption{
State-action space of aircraft handling domain, and diagrams of Follow, Chase and Land tasks.
%Handling tasks.
%. The state contains the positions, attitudes, velocities and accelerations of both the ego jet (EJ) and reference jet (RJ), and the action specifies control demands for the ego jet only. We consider three
%Match: match the fixed pose of RJ.
%Follow: turn to fly in formation with RJ on a linear path. Chase: maintain distance/line of sight to RJ as it turns randomly. Land: approach a runway using RJ as a reference.
}
\label{fig:fastjet}
\end{figure}

To formulate the aircraft handling problem as an MDP, we consider a simple set piece setup, in which the piloting agent is given a short time window (called an \textit{episode}) to manoeuvre its aircraft (the \textit{ego jet}, EJ) in a particular manner relative to a second \textit{reference jet} (RJ) whose motion, if any, is considered part of the environment dynamics.
The state space $\mathcal{S}$ contains the positions, attitudes, velocities and accelerations of both EJ and RJ, and the action space $\mathcal{A}$ consists of pitch, roll, yaw and thrust demands for EJ only.
% The dynamics function $D$ uses a basic physics engine, including gravity and air resistance.
The EJ dynamics function integrates these demands with a simplified physics engine, including gravity and air resistance. RJ dynamics, as well as the conditions of episode initialisation and termination, vary between tasks (see below).
This set piece formulation strikes a balance between simplicity and generality; many realistic scenarios faced by a fast jet pilot involve interaction with a single other airborne entity. It provides scope for the definition of many alternative tasks given the same state and action spaces, and largely unchanged dynamics.
In this work, we consider the three tasks shown in Figure \ref{fig:fastjet}:
\begin{itemize}
\item \textbf{Follow}: RJ follows a linear horizontal flight path at a constant velocity, which is oriented opposite to the initial velocity of EJ. The task is to turn onto and then maintain the path up to the episode time limit of $20$ timesteps ($\approx 20$ seconds, as timesteps are at approximately $1$Hz). This constitutes a very simple form of formation flight.
\item \textbf{Chase}: RJ follows an erratic trajectory generated by random control inputs, and the task is to chase it, maintaining distance and line of sight, without taking EJ below a safe altitude. Episodes terminate after $20$ timesteps.
\item \textbf{Land}: The task is to execute a safe approach towards landing on a runway, where RJ represents the ideal landing position (central, zero altitude, slight upward pitch). EJ is initialised at a random altitude, pitch, roll and offset, such that landing may be challenging but always physically possible. An episode terminates if EJ passes RJ along the axis of the runway, or after $25$ timesteps otherwise.
\end{itemize}

The central thesis of this paper is that there exists no unambiguous model of good aircraft handling behaviour.
For instance, experts may agree that the Chase task involves a trade-off between speed of response and smoothness of flight, but may also have nebulous and divergent definitions of these properties and their relative importance.
However, to quantitatively evaluate our method, we adopt the artificial construct of synthetic oracles based on `ground-truth' reward functions, as proxies for real human evaluators.
The oracles functions use their reward functions to provide evaluative preference feedback to our reward learning method according to a model defined in Section \ref{sec:pbrl}.
The precise nature of the oracle reward functions is secondary, and those given below are among many equally reasonable alternatives, but we dedicated several hours of development time to ensuring they incentivise reasonable behaviour upon visual inspection. The difficulty and seeming arbitrariness of such a manual reward design process is precisely why reward learning (ultimately from real human preferences) is a compelling proposition. The oracles are all defined using a common set of state-action features, which are enumerated and described in Table \ref{tab:features} (see Appendix):
\begin{itemize}
\item \textbf{Follow}: The oracle prioritises closing the distance between EJ and RJ, and matching their upward axes:
\vspace{-0.1cm}$$
R_\text{follow}=-(\texttt{dist}+0.05\times\texttt{closing}\ \texttt{speed}+10\times\texttt{up}\ \texttt{error}).
\vspace{-0.1cm}$$
\item \textbf{Chase}: The oracle prioritises keeping RJ at a distance of $20$ and within EJ's line of sight, while keeping EJ oriented upright. It also has a large penalty for dropping below a safe altitude of $50$ (\textit{note that the square brackets $[\cdot]$ are Iverson notation, which is used throughout this paper for indicator functions}):
\vspace{-0.1cm}$$
R_\text{chase}=-(\text{abs}(\texttt{dist}-20)+10\times\texttt{los}\ \texttt{error}+5\times\texttt{abs}\ \texttt{roll}+100\times[\texttt{alt}<50]).
\vspace{-0.1cm}$$
\item \textbf{Land}: The oracle for this task is the most complex, including terms that incentivise continual descent, penalise g-force and engine thrust, and punish EJ for contacting the ground before the runway $[\texttt{alt}<0.6]$:
\vspace{-0.1cm}\begin{gather*}
R_\text{land}=-(0.05\times\texttt{abs}\ \texttt{lr}\ \texttt{offset}+0.05\times\texttt{alt}
+\texttt{hdg}\ \texttt{error}+\texttt{abs}\ \texttt{roll}+0.5\times\texttt{pitch}\ \texttt{error}\\
+\ 0.25\times(\texttt{yaw}\ \texttt{rate}+\texttt{roll}\ \texttt{rate}+\texttt{pitch}\ \texttt{rate})+0.1\times\texttt{g}\ \texttt{force}+0.025\times\texttt{thrust}+0.05\times\texttt{delta}\ \texttt{thrust}\\
+\ [\texttt{delta}\ \texttt{dist}\ \texttt{hor}>0]
+2\times[\texttt{delta}\ \texttt{alt}>0]+[\texttt{abs}\ \texttt{lr}\ \texttt{offset}>10]+10\times[\texttt{alt}<0.6]).
\vspace{-0.1cm}\end{gather*}
\end{itemize}

\vspace{-0.4cm}
\section{Preference-based Reward Learning}
\label{sec:pbrl}
\vspace{0.2cm}

We adopt the preference-based approach to reward learning, in which a human (or in our case, oracle) evaluator is presented with pairs of agent trajectories (sequences of state, action, next state \textit{transitions}) and expresses which they prefer as a solution to a given task of interest.
A reward function is learnt to reconstruct the pattern of preferences.

We assume that the evaluator observes a trajectory $\xi^i$ as a sequence $\smash{(\textbf{x}^i_1,...,\textbf{x}^i_{T^i})}$, where $\smash{\textbf{x}^i_t=\phi(s^i_{t-1},a^i_{t-1},s^i_t)\in\mathbb{R}^F}$ represents a single transition as an $F$-dimensional feature vector.
The choice of feature function $\phi$ is crucial.
For our experiments, we consulted with engineers with experience in aerospace simulation and control algorithms to define $F=30$ features that are sufficiently expressive to capture the important information for all three of our target tasks, without being overly specialised to one or providing too much explicit guidance to the reward learning process.
These features are given in Table \ref{tab:features} (see Appendix). Note that all three oracles given above use a subset of the $30$ features, meaning that optimal reward learning is possible in principle but requires accurate feature selection.

\begin{figure}[!ht]
\centering
\includegraphics[width=\textwidth]{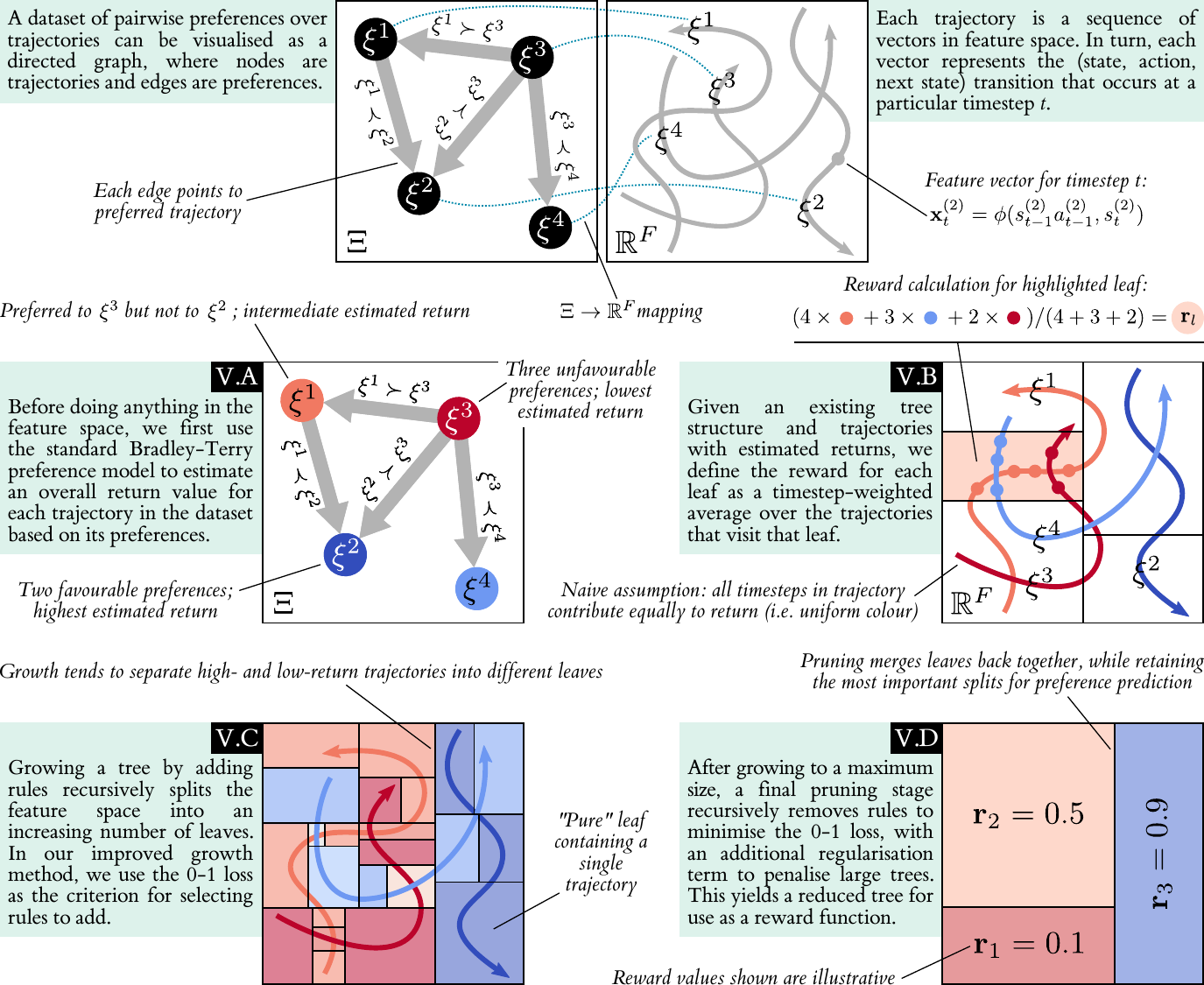}
\caption{\textit{Top}: The input to preference-based reward learning is a directed graph over a trajectory set, each of which is a sequence of points in $\mathbb{R}^F$ (blue connectors show mapping). \textit{Bottom}: 
The four stages of reward tree learning: return estimation (Section \ref{sec:traj_level}); leaf-level reward prediction (\ref{sec:reg_dataset}); tree growth (\ref{sec:growth}); pruning (\ref{sec:pruning}).
}
\label{fig:problem_and_solution}
\end{figure}

Given a set of $N$ trajectories, $\Xi=\{\xi^i\}_{i=1}^N$, the evaluator is consulted to provide $K\leq N(N-1)/2$ pairwise preference labels, $\smash{\mathcal{L}=\{(i,j)\}_{k=1}^K}$, each of which indicates that the $j$th trajectory is preferred to the $i$th (denoted by $\xi^j\succ\xi^i$).
Figure~\ref{fig:problem_and_solution} (top) shows how a preference dataset $\mathcal{D}=(\Xi,\mathcal{L})$ can be viewed as a directed graph.

% \begin{figure}
% \centering
% \includegraphics[width=\textwidth]{figures/problem_annotated.pdf}
% \caption{The input to preference-based reward learning is a directed graph over a trajectory set $\Xi$, each of which is a sequence of points in $\mathbb{R}^F$ (blue connectors show mapping).}
% \label{fig:problem_annotated}
% \end{figure}

To learn a reward function from $\mathcal{D}$, we must assume a generative preference model. Typically, it is assumed that the evaluator tends to prefer trajectories that have higher summed reward (or \textit{return}) according to a latent reward function over the feature space, $R:\mathbb{R}^F\rightarrow\mathbb{R}$, which represents their tacit understanding of the task.
However, they are liable to occasionally make mistakes in their judgement.
This is formalised by the Bradley-Terry preference model \cite{bradley1952rank}:
\vspace{-0.2cm}
\begin{equation}
    \label{eq:bradley-terry}
    P(\xi^j\succ\xi^i|R)=
    \frac{1}{1+\exp(\frac{1}{\beta}(G(\xi^i|R)-G(\xi^j|R)))},
    \vspace{-0.3cm}
\end{equation}
where $G(\xi^i|R)=\sum\nolimits_{t=1}^{T^i}R(\textbf{x}^i_t)$ and $\beta>0$ is a
% rationality
temperature
coefficient determining the probability of mistakes.

In our oracle experiments, synthetic preferences are generated according to Equation \ref{eq:bradley-terry} using the corresponding $R_\text{follow}$, $R_\text{chase}$ or $R_\text{land}$, thereby adhering to the modelling assumption.
In our main experiments, we set $\beta=0$, which results in the oracles deterministically selecting trajectories with higher return (ties broken uniform-randomly). We subsequently study cases with $\beta>0$, which provide a more realistic emulation of real human preference data.

Given a preference dataset and an assumed generative model such as Equation \ref{eq:bradley-terry}, the objective of reward learning is to approximate the latent reward function $R$ within some learnable function class $\mathcal{R}$. This problem is often formalised as minimising the negative log-likelihood (NLL) loss over the preferences in $\mathcal{L}$ \cite{christiano2017deep,lee2021pebble}.
% https://towardsdatascience.com/cross-entropy-negative-log-likelihood-and-all-that-jazz-47a95bd2e81#:~:text=Negative%20log%2Dlikelihood%20minimization%20is,up%20the%20correct%20log%20probabilities.%E2%80%9D
\citet{wirth2016model} also use the discrete $0$-$1$ loss, which considers only the directions of predicted preferences rather than their strengths.
Our model uses both of these losses at different stages in its learning process. They are respectively defined as follows:
\vspace{-1cm}
\begin{multicols}{2}
  \begin{equation}
    \label{eq:nll_loss}
    \ell_{\text{NLL}}(\mathcal{D},R)=
    \sum_{(i,j)\in\mathcal{L}}-\log P(\xi^j\succ\xi^i|R);
  \end{equation}\break
  \begin{equation}
    \label{eq:0_1_loss}
    \ell_{\text{0-1}}(\mathcal{D},R)=
    \sum_{(i,j)\in\mathcal{L}}[
    %G(\xi^j|R)\geq G(\xi^i|R)
    P(\xi^j\succ \xi^i|R) \leq 0.5
    ].
  \end{equation}
\end{multicols}
\vspace{-0.2cm}

In early prior work, $\mathcal{R}$ was often the class of linear models $R(\textbf{x})=\textbf{w}^\top\textbf{x}$ \citep{sadigh2017active}, which are easy to interpret but have limited expressiveness, so cannot scale to complex tasks.
More recently, it has been common to use deep NNs \citep{christiano2017deep} (or multi-network ensembles thereof), which are far more powerful, while remaining tractably learnable by gradient-based optimisation.
However, the complex multilayered architecture of a deep NN resists human scrutiny and interpretation.
This motivates a `third way' function class that compromises between these two extremes.

\vspace{-0.4cm}
\section{Interpretable Reward Learning with Trees}
\label{sec:model_induction}
\vspace{0.2cm}

As an intermediate option between the limited expressiveness of linear models and the uninterpretable complexity of NNs, our recent prior work proposes reward trees \cite{bewley2022interpretable}. Here, $\mathcal{R}$ is the class of axis-aligned decision trees, which are hierarchies of local rules constructed from the feature set.
Reward trees admit visual and textual representation, and produce a traceable decision path for each reward prediction.
While methods exist for learning trees by gradient-based optimisation \citep{suarez1999globally}, these are of the oblique (c.f. axis-aligned) kind, whose multi-feature rules are far harder to interpret in high dimensions. Therefore, instead of optimising one of the losses from Equations \ref{eq:nll_loss} and \ref{eq:0_1_loss} end-to-end, we use a
multi-stage induction method with a proxy objective at each stage.
The four stages outlined below, and depicted in Figure \ref{fig:problem_and_solution}, depart from our original method in several respects, which we highlight and justify where relevant.

\vspace{-0.4cm}
\subsection{Trajectory-level Return Estimation}
\label{sec:traj_level}
\vspace{0.2cm}

This first stage of our method considers the $N$ trajectories as atomic units, and uses the preference graph to construct a vector of return estimates $\textbf{g}\in\mathbb{R}^N$, which should be higher for more preferred trajectories (blue in Figure \ref{fig:problem_and_solution} (V.A), c.f. red). This is a vanilla preference-based ranking problem,
%of the kind routinely faced in AI, psychology and economics,
and admits a standard solution.
In \cite{bewley2022interpretable}, we use a least squares matrix method to solve for $\textbf{g}$ under Thurstone's Case V preference model \citep{gulliksen1956least}. For consistency with prior work, and to avoid an awkward clipping step which biases preference probabilities to enable matrix inversion, we now use a gradient method to minimise $\ell_{\text{NLL}}$ under the Bradley-Terry model instead. Concretely, the objective for this stage is
\vspace{-0.2cm}
\begin{equation}
    \label{eq:proxy_loss_trajectory_level}
    \underset{\textbf{g}\in\mathbb{R}^N}{\text{argmin}}\Big[\sum_{(i,j)\in\mathcal{L}}-\log \frac{1}{1+\exp(\textbf{g}^i-\textbf{g}^j)}\Big],
    \vspace{-0.2cm}
\end{equation}
which we optimise by gradient descent using the Adam optimiser \cite{kingma2014adam}.
As a minor detail, we post-normalise $\textbf{g}$ so that its elements have a standard deviation equal to the mean trajectory length in $\Xi$, $\sum_{i=1}^N T^i/N$, and a common sign (i.e. all positive or negative). We anecdotally find that this aids the interpretability of the final reward tree.

\vspace{-0.4cm}
\subsection{Leaf-level Reward Prediction}
\label{sec:reg_dataset}
\vspace{0.2cm}

The vector $\textbf{g}$ estimates trajectory returns, but the ultimate aim of reward learning is to decompose these into sums of rewards for the constituent transitions, then generalise to make reward predictions for unseen data (e.g. novel trajectories executed by an RL agent). Our core contribution is to do this using a tree model $\mathcal{T}$, consisting of a hierarchy of rules that partition the transition-level feature space $\mathbb{R}^F$ into $L_\mathcal{T}$ hyperrectangular regions called \textit{leaves}.
Given such a tree, each leaf $l\in\{1..L_\mathcal{T}\}$ is associated with a reward prediction $\textbf{r}_l$ as follows. Let the function $\text{leaf}_\mathcal{T}:\mathbb{R}^F\rightarrow\{1..L_\mathcal{T}\}$ map a feature vector $\textbf{x}\in\mathbb{R}^F$ to the leaf in which it resides by propagating it through the rule hierarchy.
%Next, let $\tau\in\{1\sdot T^\text{total}\}$ denote the indexing variable for elements of $\textbf{X}$ and $\textbf{r}$, and $\mathcal{T}(l)=\{\tau:\text{leaf}(\textbf{X}_\tau)=l\}$ denote the indices that lie in leaf $l$.
$\textbf{r}_l$ is defined as an average over $\textbf{g}$, weighted by the proportion of time that each trajectory in $\Xi$ spends in $l$:
\vspace{-0.25cm}
\begin{equation}
    \label{eq:reward_prediction}
    \textbf{r}_l=\sum_{i=1}^N\frac{\textbf{g}^i}{T^i}\frac{\sum_{t=1}^{T^i}[\text{leaf}_\mathcal{T}(\textbf{x}^i_t)=l]}{\sum_{j=1}^N\sum_{t=1}^{T^j}[\text{leaf}_\mathcal{T}(\textbf{x}^j_t)=l]}.
    \vspace{-0.15cm}
\end{equation}
%In matrix form, $\textbf{r}=\textbf{V}(\textbf{g}\oslash\textbf{t})=\textbf{V}(((\textbf{A}^\top\textbf{A})^{-1}\textbf{A}^\top\textbf{y})\oslash\textbf{t})$
The effect of Equation \ref{eq:reward_prediction} is to assign higher reward to leaves that contain more timesteps from trajectories with high $\textbf{g}$ values (i.e. those more commonly preferred in the preference dataset).
While ostensibly na{\"i}ve, we find that this time-weighted credit assignment is more robust than several more complex alternatives. It reduces the free parameters in subsequent induction stages, permits fast implementation, and provides an intuitive interpretation of predicted reward that is traceable back to a $
\textbf{g}$ value and timestep count for each $\xi^i\in\Xi$. Figure~\ref{fig:problem_and_solution} (V.B) shows how $4$, $2$ and $3$ timesteps from $\xi^1$, $\xi^3$ and $\xi^4$ are averaged over to yield the reward prediction for one leaf (indicated by the orange shading).

This definition provides the basis for using $\mathcal{T}$ as a reward function. Given an arbitrary feature vector $\textbf{x}\in\mathbb{R}^F$, we simply look up the reward of the leaf in which it resides:
\vspace{-0.2cm}
\begin{equation}
    R_\mathcal{T}(\textbf{x})=\textbf{r}_{\text{leaf}_\mathcal{T}(\textbf{x})}.
\end{equation}

\vspace{-0.8cm}
\subsection{Tree Growth}
\label{sec:growth}
\vspace{0.2cm}

Recall that the objective of preference-based reward learning is to adjust the parameters of the reward model in order to minimise some loss function over $\mathcal{D}$, such as those in Equations \ref{eq:nll_loss} and \ref{eq:0_1_loss}. When the model is a tree, this is achieved by the discrete operations of growth (adding partitioning rules to increase the number of leaves) and pruning (removing rules to decrease the number of leaves). Given a tree $\mathcal{T}$, a new rule has the effect of splitting the $l$th leaf with a hyperplane at a location $c\in\mathcal{C}_f$ along the $f$th feature dimension (where $\mathcal{C}_f\subset\mathbb{R}$ is a set of candidate split thresholds, e.g. all midpoints between unique values in $\Xi$). Let $\mathcal{T}+[lfc]$ denote the newly-enlarged tree. Splitting recursively creates an increasingly fine partition of $\mathbb{R}^F$. Figure \ref{fig:problem_and_solution} (V.C) shows an example of a reward tree with 23 leaves.

A central issue is the criterion for selecting the next rule to add.
In \cite{bewley2022interpretable}, we use the proxy objective of minimising the local variance of $\textbf{g}$ values in each leaf, which exactly corresponds to the classic CART algorithm \citep{breiman2017classification}. While very fast to compute, this criterion is only loosely aligned with the reconstruction of the preferences in $\mathcal{D}$.
In the present work, we propose and investigate the more direct criterion of greedily minimising the $\ell_\text{0-1}$ of the enlarged tree $\mathcal{T}+[lfc]$:
\vspace{-0.25cm}
\begin{equation}
    \label{eq:split_criterion}
    \text{argmin}_{1\leq l\leq L_\mathcal{T},\ 1\leq f\leq F,\ c\in\mathcal{C}_f}\ \left[\ \ell_{\text{0-1}}(\mathcal{D},R_{\mathcal{T}+[lfc]})\ \right],
    \vspace{-0.15cm}
\end{equation}
i.e. selecting splits to minimise the number of incorrectly-predicted preferences.
In Section \ref{sec:results}, we show that switching to this criterion consistently improves reward learning and agent policy performance on the aircraft handling tasks.
%(we also tried a criterion based on $\ell_\text{NLL}$, but found it to be far more computationally costly and prone to overfitting).
% Implementing it efficiently required a major reformulation of the tree growth algorithm; we provide vectorised, just-in-time compiled code for this in the Supplementary Material.
Recursive splitting stops when $\ell_\text{0-1}$ cannot be reduced by any single split, or a tree size limit $L_\mathcal{T}=L_\text{max}$ is reached.
%Quick aside about the feature function $\phi$: \citet{tien2022study} find that careful feature design reduces causal confusion in PbRL. Trees inherently perform feature selection, so can start with a large bank of candidates.
%\citet{katz2021preference} learns features by gradient descent offline after preference collection

\vspace{-0.4cm}
\subsection{Tree Pruning}
\label{sec:pruning}
\vspace{0.2cm}

Growth is followed by a pruning stage which reduces the size of the tree by rule removal. This is beneficial for both performance (\citet{tien2022study} find that limiting model capacity lowers the risk of causal confusion in preference-based reward learning) and human comprehension (in the language of \citet{jenner2022preprocessing}, pruning is a form of \textit{``processing for interpretability"}). Given a tree $\mathcal{T}$, one pruning operation has the effect of merging two leaves into one by removing the rule at the common parent node. Let $\mathbb{T}$ denote the sequence of nested subtrees induced by pruning the tree recursively back to its root, at each step removing the rule that minimises the next subtree's $\ell_{\text{0-1}}$.
%The final step of model induction is to
We select the $\mathcal{T}\in\mathbb{T}$ that minimises $\ell_{\text{0-1}}$, additionally regularised by a term that encourages small trees:
\vspace{-0.25cm}
\begin{equation}
    \text{argmin}_{\mathcal{T}\in\mathbb{T}}[\ell_{\text{0-1}}(\mathcal{D},R_\mathcal{T})+\alpha L_\mathcal{T}],
    \vspace{-0.15cm}
\end{equation}
where $\alpha\geq 0$ is a regularisation coefficient.
% Note that even with $\alpha=0$ pruning may still yield a reduced tree, as unlike in traditional decision tree induction, the effect of individual rules on $\ell_{\text{0-1}}$ depends on the order in which they are added or removed.
In the example in Figure \ref{fig:problem_and_solution} (V.D), pruning yields a final tree with $3$ leaves, for which illustrative leaf-level reward predictions are shown.
%Also tried growing multiple trees with different random partitions of the preference graph for growth and pruning and maintaining a ``forest" of past trees but didn't warrant additional complexity.
% Such a tree can be used as a reward function for training an RL agent to satisfy the \N{human}'s preferences.

\vspace{-0.4cm}
\section{Online Learning Process}
\label{sec:online_learning}
\vspace{0.2cm}

\vspace{-0.4cm}
\subsection{Iterated Policy and Reward Learning}
\vspace{0.2cm}

Sections \ref{sec:pbrl} and \ref{sec:model_induction} do not discuss the origins of the trajectories $\Xi$, or how reward tree learning should be integrated with the process of policy learning by RL. Following most recent work since \citet{christiano2017deep}, we resolve both questions with an iterative bootstrapped approach, in which the current reward tree defines the RL agent's reward function at each point in learning.
During learning episode $i$, the agent uses its latest policy to produce a new trajectory $\xi^i$. We immediately connect $\xi^i$ to the preference graph by asking the human (read: oracle) to compare it to $K_\text{batch}$ random trajectories from the existing set. We then update the reward tree on the full preference graph via the stages in Section~\ref{sec:model_induction}. We find that our original method of starting growth from the current state of the tree causes lock-in to poor initial solutions,
so instead re-grow from scratch (i.e. starting from a single leaf) on each update. The rule structure nonetheless tends to stabilise, as the enlarging preference graph becomes increasingly similar for later updates.
For the $(i+1)$th episode, the RL agent then attempts to optimise its policy with respect to the newly-updated reward tree. 
By iterating this process up to a total preference budget $K_\text{max}$ and/or episode budget $N_\text{max}$,
we hope to converge to both a reward
tree that reflects the human's preferences, and an agent policy that satisfies those preferences.

\vspace{-0.4cm}
\subsection{Model-based RL Algorithm}
\label{sec:model_based}
\vspace{0.2cm}

Online reward learning is generally agnostic to the structure of the policy learning agent; this modularity is hailed as an advantage over other human-agent teaching paradigms \citep{leike2018scalable}. Following most recent work, in \cite{bewley2022interpretable} we use a model-free RL agent, specifically soft actor-critic (SAC) \citep{haarnoja2018soft}. However, other work \citep{reddy2020learning,rahtz2022safe} uses model-based RL (MBRL) agents that leverage learnt dynamics models and planning. MBRL is attractive in the reward learning context because it disentangles the predictive and normative aspects of decision-making. Since (assuming no changes to the environment) dynamics remain stationary during online reward learning, the amount of re-learning required is reduced.
MBRL can also be very data-efficient; our preliminary experiments found that switching from SAC to a model-based algorithm called PETS \citep{chua2018deep} reduces environment interaction during reward learning by orders of magnitude. PETS selects actions by decision-time planning through a learnt dynamics model $D':\mathcal{S}\times\mathcal{A}\rightarrow\Delta(\mathcal{S})$ up to a horizon $H$. In state $s$, planning searches for a sequence of $H$ future actions that maximise return according to the current reward tree:
\vspace{-0.2cm}
\begin{equation}
    \label{eq:pets_action_sequence_eval}
    \underset{(a_0,...,a_{H-1})\in\mathcal{A}^H}{\text{argmax}}\mathbb{E}_{D'}\Big[\sum\nolimits_{h=0}^{H-1}\gamma^h R_\mathcal{T}(\phi(s_h,a_h,s_{h+1}))\Big],\ \text{where}\ s_0=s,\ s_{h+1}\sim D'(s_h,a_h).
    \vspace{-0.2cm}
\end{equation}
The first action $a=a_0$ is executed, and then the agent re-plans on the next timestep.
%This is an example of model predictive control.
In practice, $D'$ is an ensemble of probabilistic NNs, the expectation over $D'$ is replaced by a Monte Carlo estimate, and the optimisation is approximated by the iterative cross-entropy method.

We use PETS agents in all experiments in this paper.
In our implementation, $D'$ is an ensemble of five NNs, each with four hidden layers of $200$ units.
% State vectors are pre-normalised by applying a hand-specified scale factor to each dimension.
Planning operates over a time horizon of $H=10$, with no discounting ($\gamma=1$), and consists of $10$ iterations of the cross-entropy method. Each iteration samples $20$ candidate action sequences from an independent Gaussian, of which the top five in terms of return are identified as \textit{elites}, then updates the sampling Gaussian towards the elites with a learning rate of $0.5$.
This parameter affects how rapidly the distribution narrows towards a deterministic action sequence. In turn, this determines whether the agent only exploits actions that appear optimal under the current reward model $R_\mathcal{T}$, or explores more diverse behaviour that might be preferred by the human evaluator. Correctly balancing the explore/exploit trade-off is especially crucial in the reward learning context.

We find that the particular dynamics of the aircraft handling environment permit us to pre-train $D'$ on random data, and accurately generalise to states encountered during online reward learning. This means we perform no further updates to $D'$ while reward learning is ongoing. As well as improving execution speed, this avoids complexity and convergence issues arising from having two interacting learning processes (note that simultaneous learning is unavoidable with model-free RL). To pre-train, we collect $1e^5$ transitions using a uniform random policy, then update each network on $1e^5$ independently sampled mini-batches of $256$ transitions, using the mean squared error loss over next-state predictions.

\vspace{-0.4cm}
\subsection{Diagram of Online Learning Process}
\vspace{0.2cm}

\begin{figure}[h!]
\centering
\includegraphics[width=\textwidth]{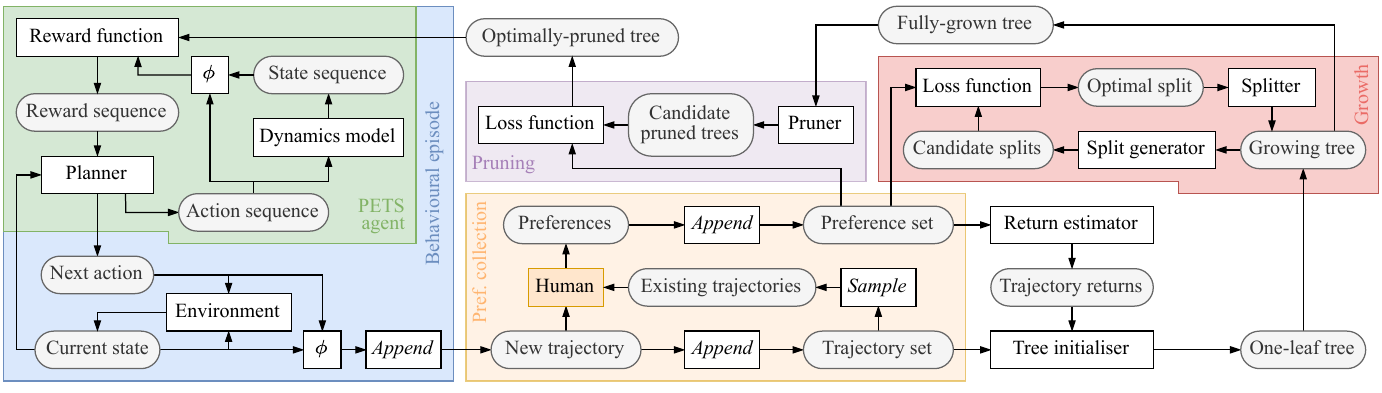}
\vspace{-0.6cm}
\caption{Online preference-based reward tree learning with model-based PETS agents.}
\label{fig:online_learning_diagram}
\vspace{-0.7cm}
\end{figure}

Figure~\ref{fig:online_learning_diagram} summarises the learning approach taken in this paper, with the details of some steps omitted for clarity. Although we obtain preferences from synthetic oracles, the process using a real human evaluator would be identical.

\vspace{-0.4cm}
% \section{Experimental  Scope and Domain}
\section{Experiments and Results}
\label{sec:results}
\vspace{0.2cm}

In this section, we combine quantitative and qualitative evaluations to assess the performance of reward tree learning with oracle preferences on the aircraft handling tasks, specifically in comparison to the de facto standard approach of using NNs. We also illustrate how the intrinsic interpretability of reward trees allows us to analyse what they have learnt.

In all experiments, we use the following set of hyperparameters for tree learning. These were identified through informal search, and we make no claim of optimality, but they do lead to acceptable performance on the three tasks of varying complexity. The fact that we did not need to invest significant time in tuning indicates a general insensitivity of the method to precise hyperparameter values, which is widely seen as practically advantageous.
\begin{itemize}
    \item Trajectory return estimation using the Adam optimiser with a learning rate of $0.1$. Optimisation is stopped when the loss $\ell_{\text{NLL}}$ changes by $<1e^{-5}$ between successive gradient steps.
    \item Per-feature candidate split thresholds $\mathcal{C}_f$ defined as all midpoints between adjacent unique values in the trajectory set $\Xi$. These are recomputed on each update.
    \item Tree size limit $L_\text{max}=100$.
    \item Tree size regularisation coefficient $\alpha=5e^{-3}$.
\end{itemize}

In constructing the NN baseline, we aimed to retain as much of the algorithm structure from Figure~\ref{fig:online_learning_diagram} as possible, so that only the model architecture varies. The result is that we perform policy learning with PETS, trajectory pair sampling and oracle preference collection identically, and update the reward model after every episode.
However, in place of the multi-stage tree growth and pruning process, we perform model updates by mini-batch gradient descent with respect to $\ell_{\text{NLL}}$ from Equation \ref{eq:nll_loss}.
In all experiments, we follow \citet{lee2021pebble} in implementing the NN reward model as a three-layer network with 256 hidden units each and leaky ReLU activations, and performing the gradient-based updates using the Adam optimiser \citep{kingma2014adam} with a learning rate of $3e^{-4}$. On each update, we sample $M=100$ mini-batches of size $B=32$ and take one gradient step per mini-batch.

\vspace{-0.4cm}
\subsection{Quantitative Performance}
\label{sec:fid_perf_results}
\vspace{0.2cm}

In our main experiments, we use ideal, error-free oracles, which provide preferences according to Equation \ref{eq:bradley-terry} with $\beta=0$.
We evaluate online reward learning with PETS using trees with the $\ell_{\text{0-1}}$ split criterion, baselined against our original variance criterion, as well as the de facto standard
of NN reward learning.
%via gradient descent on $\ell_{\text{NLL}}$
We use $K_\text{max}=1000$ preferences over $N_\text{max}=200$ online trajectories (i.e. those generated by the PETS agents during learning) and run $10$ repeats.

As a headline statistic, Table \ref{tab:orr} reports the \textit{oracle regret ratio} (ORR): the drop in average oracle return of PETS agents deployed using each trained reward model compared with directly using the oracle reward function, as a fraction of the drop to a random policy (lower is better). 
This gives a normalised measure of how well reward learning performs compared with the ideal case of direct access to the true reward.
We report the median and minimum ORR values across the $10$ repeats for each task-model pairing.
%as in \citet{lindner2021information}
We observe that:
1) NN reward learning is strong on all tasks,
2) switching to a tree induces a small but variable performance hit,
3) $\ell_{\text{0-1}}$ splitting outperforms the variance-based method, and 4) both NN and tree models sometimes exceed the direct use of the oracle for policy learning (negative ORR). This counter-intuitive phenomenon has been observed before \citep{cao2021weak} and may be due to improved shaping in the learnt reward.

\vspace{-0.1cm}
\setlength{\tabcolsep}{0.49em}
\begin{table}[h!]
\renewcommand{\arraystretch}{1.2}
\centering
\small
\begin{tabular}{cc|c|c|c|c|c|c|c|c}
\cline{2-10}
\textbf{Table 1\quad Median (top)} & 
\multicolumn{3}{c|}{Follow}    & \multicolumn{3}{c|}{Chase}    & \multicolumn{3}{c}{Land}     \\
 \textbf{and minimum (bottom)} & NN & Tree(0-1) & Tree (var) & NN & Tree (0-1) & Tree (var) & NN & Tree (0-1) & Tree (var) \\
\cline{2-10}
 \textbf{oracle regret ratios for} & $.000$ & $.120$ & $.284$ & $-.030$ & $.040$ & $.126$ & $.014$ & $.050$ & $.062$ \\
\textbf{all tasks and models.} & $-.010$ & $.057$ & $.158$ & $-.051$ & $-.011$ & $.065$ & $-.030$ & $.011$ & $.010$\\
\cline{2-10}
\end{tabular}
\fakecaption
\label{tab:orr}
\vspace{-0.3cm}
\end{table}

\begin{figure}
\centering
\includegraphics[width=\textwidth]{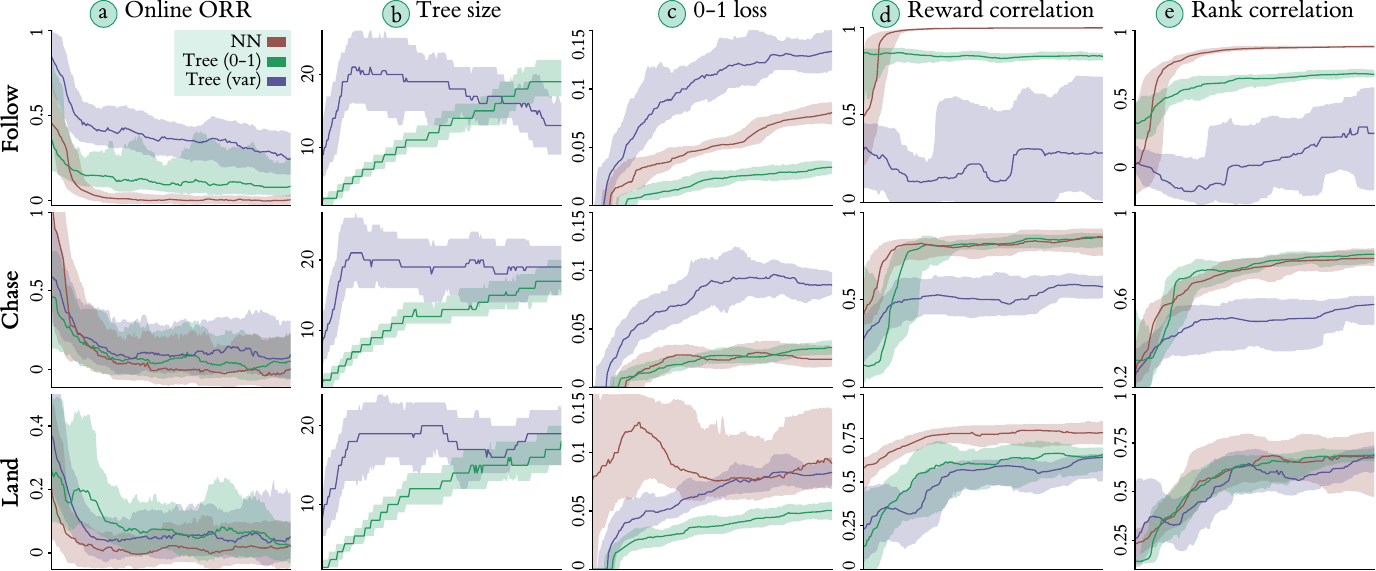}
\caption{Time series of metrics for online NN- and tree-based reward learning on all three tasks.\vspace{-0.165cm}}
\label{fig:results_main}
\end{figure}

Figure \ref{fig:results_main} expands these results with more metrics, revealing learning trends not captured by headline ORR values. Metrics are plotted as time series over the $200$ learning episodes (sliding-window medians and interquartile ranges (IQRs) across repeats). In the left column (\textbf{a}), the ORR of online trajectories shows how agent performance converges. \textbf{For Follow, there is a gap between the models}, with $\ell_{\text{0-1}}$ splitting visibly aiding performance but still lagging behind the NNs. \textbf{The learning curves for Chase and Land are more homogeneous}, and the NNs reach only slightly lower asymptotes, with overlapping IQRs.
The majority of models converge to their asymptotic performance well within the $200$ episode learning period; this learning speed is made possible by the use of model-based PETS agents.

For the reward tree models, (\textbf{b}) shows how the number of leaves changes over time.
There is notable consistency in these trends between the repeated runs.
The variance-based trees tend to grow rapidly initially before stabilising or shrinking, while \textbf{the $\ell_{\text{0-1}}$ trees enlarge more conservatively, suggesting this method is less liable to overfit} to small preference datasets. Trees of a readily-interpretable size ($\approx 20$ leaves) are produced for all tasks; it is possible that performance could be improved by independently tuning the size regulariser $\alpha$ per task.

(\textbf{c}) shows how the discrete preference loss $\ell_{\text{0-1}}$ changes over time, which tends to increase as the growing preference graph presents a harder reconstruction problem, though the shapes of all curves suggest convergence (note that random prediction gives $\ell_{\text{0-1}}=0.5$). For Follow and Land, \textbf{the trees that directly split on $\ell_{\text{0-1}}$ actually achieve lower loss than the NNs; they more accurately predict the direction of preferences in the graph}. This is an encouraging result, indicating that our models perform well on the metric for which they are directly optimised, but the fact that this does not translate into lower ORR indicates that the problems of learning a good policy and replicating the preference dataset are not perfectly correlated. This subtle point has previously been made by \citet{lindner2021information}.

\citet{gleave2021quantifying} recently highlighted the importance of comparing and evaluating learnt reward functions in a policy-invariant manner, by using a common evaluation dataset rather than on-policy data generated by agents optimising for each reward. We report such a comparison in the final two columns, where the evaluation datasets are generated by PETS agents using the oracle reward functions, with added action randomisation to increase diversity. Using these datasets, we correlate each reward model's predictions with its respective oracle at each point in learning, in terms of both transition-level rewards (\textbf{d}) and the ordinal ranking of trajectories by return (\textbf{e}), the latter via the \citet{kendall1938new} $\tau$ coefficient. The curves subtly differ, indicating that it is possible to reconstruct trajectory rankings (and by implication, any pairwise preferences) to a given accuracy with varying fidelity at the individual reward level. However, the common overall trend is that \textbf{$\ell_{\text{0-1}}$-based trees outperform variance-based ones, with NNs sometimes improving again by a smaller margin, and sometimes bringing no added benefit}. Moving top-to-bottom down the tasks, the gap between models reduces from both sides; NN performance worsens while variance-based trees improve.

% Key message: NN $>$ 0-1 $>$ var, more going on than headline figure

% reduce or even vanish moving top-to-bottom down the tasks. 
% Also show online as dotted lines, differences indicate that all (?) model types tend to overfit variance worst
% Don't need very high correlations to get low ORR

%Results support conclusion that fidelity is not the only thing to focus on, performance is distinct  \citet{lindner2021information}. For ``forgiving" tasks, can get away with poor fidelity, including significant overfitting, without much impact on performance. Huge set of roughly equivalent rewards for performance. A general question about reward learning: if can get away with lower fidelity, can get away with simpler and more structured reward functions. If performance isn't that fidelity-dependent: suggests that reward functions may be a better place to inject tree representations than policies. But now need analyse robustness to a more realistic labelling context, and combine with interpretability analysis to really demonstrate that trees are better (not just \textit{a little bit worse}).

A potentially important factor in these experiments is that the oracle reward for Follow is a linear function,
%(which is not naturally modelled by a hierarchy of discrete thresholds),
while the others contain progressively more terms and discontinuities (see Section~\ref{sec:env_and_tasks}). A trend suggested by these results is thus that \textbf{the performance gap between NNs and reward trees} (on both ORR and correlation metrics) \textbf{reduces as the true reward becomes more complex and nonlinear}. Further experiments would be needed to test this hypothesis.

\vspace{-0.4cm}
\subsection{Visual Trajectory Inspection}
\vspace{0.2cm}

\begin{figure}
\centering
\includegraphics[width=\textwidth]{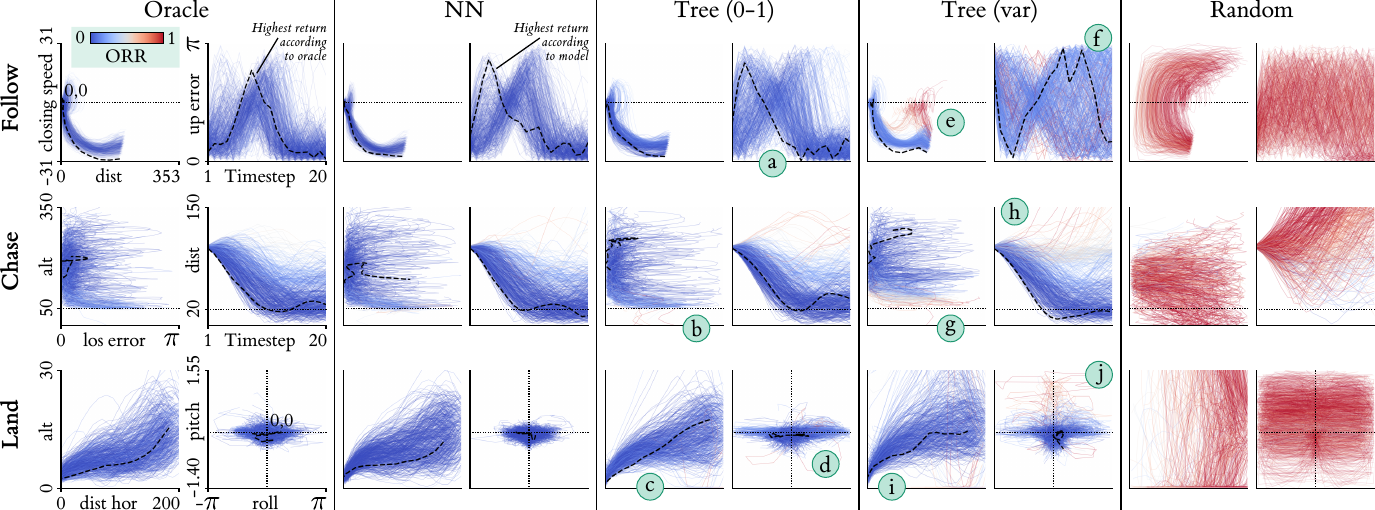}
\caption{Agent trajectories using the best models by ORR, with oracle and random for comparison.}
\label{fig:trajectory_visualisation}
\end{figure}

While useful for benchmarking, quantitative metrics provide little insight into the structure of the learnt solutions. They would also mostly be undefined when learning from real humans since the ground truth reward is unknown. We therefore complement them with a visual analysis of induced agent behaviour. Figure \ref{fig:trajectory_visualisation} plots $500$ trajectories of PETS agents using the best repeat by ORR for each task-model combination, across a range of 
%2D projections in feature-time space
features as well as time
(see Table \ref{tab:features} for a reminder of feature definitions).
Each trajectory is coloured on a red-blue scale according to its ORR.
Dashed black curves indicate the single trajectory with the highest predicted return according to each model. We also show trajectories for PETS agents with direct oracle access, which serve as the benchmark behaviour that we aim to match via reward learning, and for random policies, which perform very poorly on all three tasks.

The high-level trend is that all models are far closer to the oracle than random, with few examples of obviously unstable handling behaviour or task failure (highlighted in red, due to colouring by ORR). \textbf{While the NNs induce trajectories that are almost indistinguishable from the oracle, the $\ell_{\text{0-1}}$-based reward trees lag not far behind. The variance-based trees exhibit more anomalies.} Successes of the $\ell_{\text{0-1}}$ trees include the execution of Follow with a single banked turn before straightening up, as shown by the \texttt{up} \texttt{error} time series (\textbf{a}), where \texttt{up} $\texttt{error}=0$ is level flight.
Interestingly, both this tree and the NN reward model appear to favour a somewhat earlier turn than the oracle (i.e. peak shifted to the left on this plot).
Indeed, the trajectories for the $\ell_{\text{0-1}}$ tree are almost imperceptibly different from those of the NN, despite their quantitative performance (e.g. ORR) differing. This underlines the importance of joint quantitative-qualitative evaluation. For Chase (\textbf{b}), the $\ell_{\text{0-1}}$ tree has clearly learnt the most safety-critical aspect of the task, which is to keep the agent above the altitude threshold $\texttt{alt}<50$, below which the oracle reward is strongly negative. The threshold is violated in only eight of $500$ trajectories ($1.6\%$). 
Further evidence that the altitude threshold has been learnt correctly is discussed in Section~\ref{sec:tree_analysis}.
For Land, the $\ell_{\text{0-1}}$ tree replicates the oracle in producing a gradual reduction in \texttt{alt} (\textbf{c}) while usually keeping \texttt{pitch} close to $0$ (\textbf{d}), although the spread of \texttt{roll} values is somewhat wider.

In contrast, the agent using the variance-based tree for Follow sometimes fails to reach the target position (\textbf{e}; red trajectories), and also does not reliably straighten up to reduce \texttt{up} \texttt{error} (\textbf{f}). For Chase, the altitude threshold does not appear to have been learnt precisely, and lower-altitude trajectories often fail to close the distance to RJ (\textbf{g} and \textbf{h}; red trajectories). For Land, the variance-based tree gives a later and less smooth descent (\textbf{i}), and less consistent pitch control (\textbf{j}), than the NN or $\ell_{\text{0-1}}$-based tree, although all models produce a somewhat higher altitude profile than the oracle.

\vspace{-0.4cm}
\subsection{Sensitivity Analysis}
\label{sec:sensitivity_analysis}
\vspace{0.2cm}

\begin{figure}
\centering
\includegraphics[width=\textwidth]{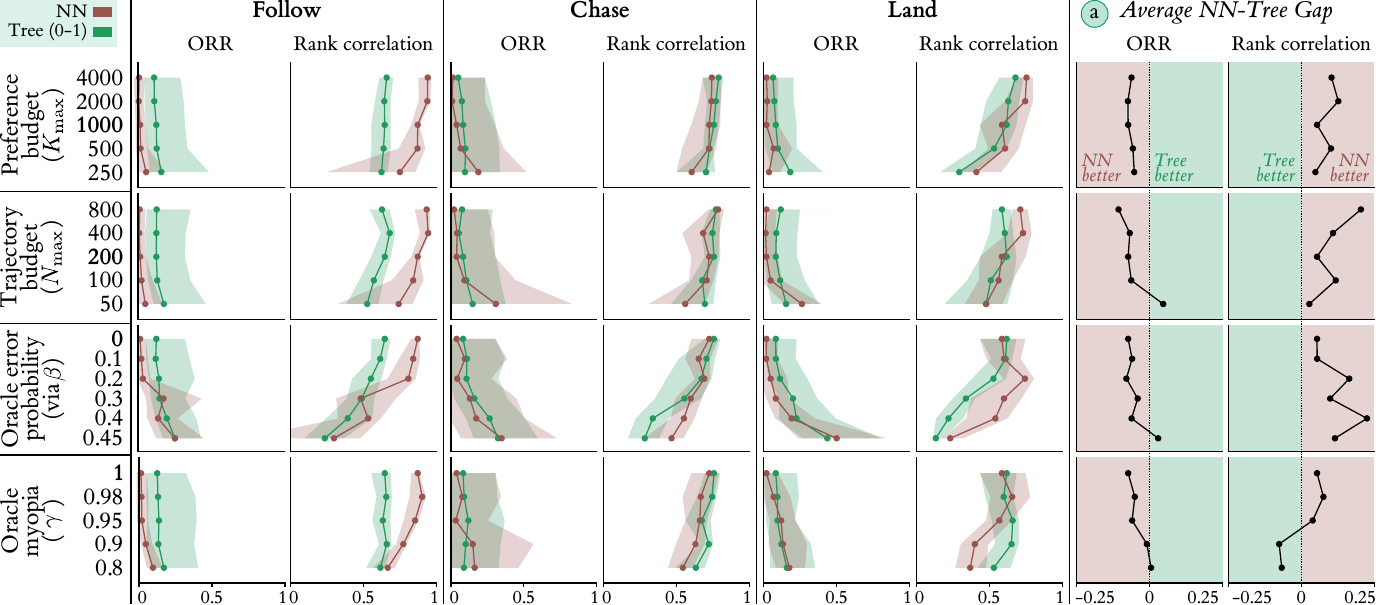}
\caption{Comparative sensitivity analysis of reward learning with NNs and trees.}
\label{fig:sensitivity_analysis}
\end{figure}

It is important to consider how learning performance degrades with reduced or corrupted data. In Figure \ref{fig:sensitivity_analysis}, we evaluate the effect of varying the number of preferences $K_\text{max}$ (with fixed $N_\text{max}=200$) and trajectories $N_\text{max}$ (with fixed $K_\text{max}=1000$) on reward learning with NNs and $\ell_{\text{0-1}}$-splitting trees. Following \citet{lee2021bpref}, we also create more human-like preference data via two modes of oracle \textit{irrationality}: preference noise (by using a nonzero Bradley-Terry temperature $\beta$ to give a desired error rate on the coverage datasets) and a myopic recency bias (by exponentially discounting earlier timesteps when evaluating trajectory returns). We run five repeats for all cases, and report the medians and interquartile ranges of ORR (lower is better) and rank correlation (closer to $1$ is better).

Both NN and tree models exhibit good robustness with respect to all four parameters. Although NNs remain superior in most cases, the gap varies, and is often reduced compared to the base cases (bold labels). \textbf{The budget sensitivity is low}, with little improvement for $K_\text{max}>1000$ and $N_\text{max}>200$, and no major drop even with $25\%$ of the data as the base case. For all tasks, \textbf{the oracle error probability can increase to around $20\%$ before significant drops in performance are observed}. These are promising indicators of the transferability of reward tree learning to limited and imperfect human data. Another general observation is that the trends for trees are somewhat smoother than for NNs, with fewer sharp jumps and fewer instances of very high spread across the five repeats.

In the right column (\textbf{a}), we summarise these results by taking the difference between the NN and tree metrics, and averaging across the three tasks. In all cases aside from rank correlation with $\beta>0$, the NN-tree gap tends to become more favourable to the trees as the varied parameter becomes more challenging (top-to-bottom). This sensitivity analysis thus indicates that \textbf{reward trees are at least as robust to difficult learning scenarios as NNs, and may even be slightly more so}.
This is another promising result for the viability of reward tree learning from real human preferences.

% \section{Supporting the \textit{Should} Claim: ... Explainability}
\vspace{-0.4cm}
\subsection{Tree Structure Analysis}
\label{sec:tree_analysis}
\vspace{0.2cm}

\begin{figure}
\centering
\includegraphics[width=\textwidth]{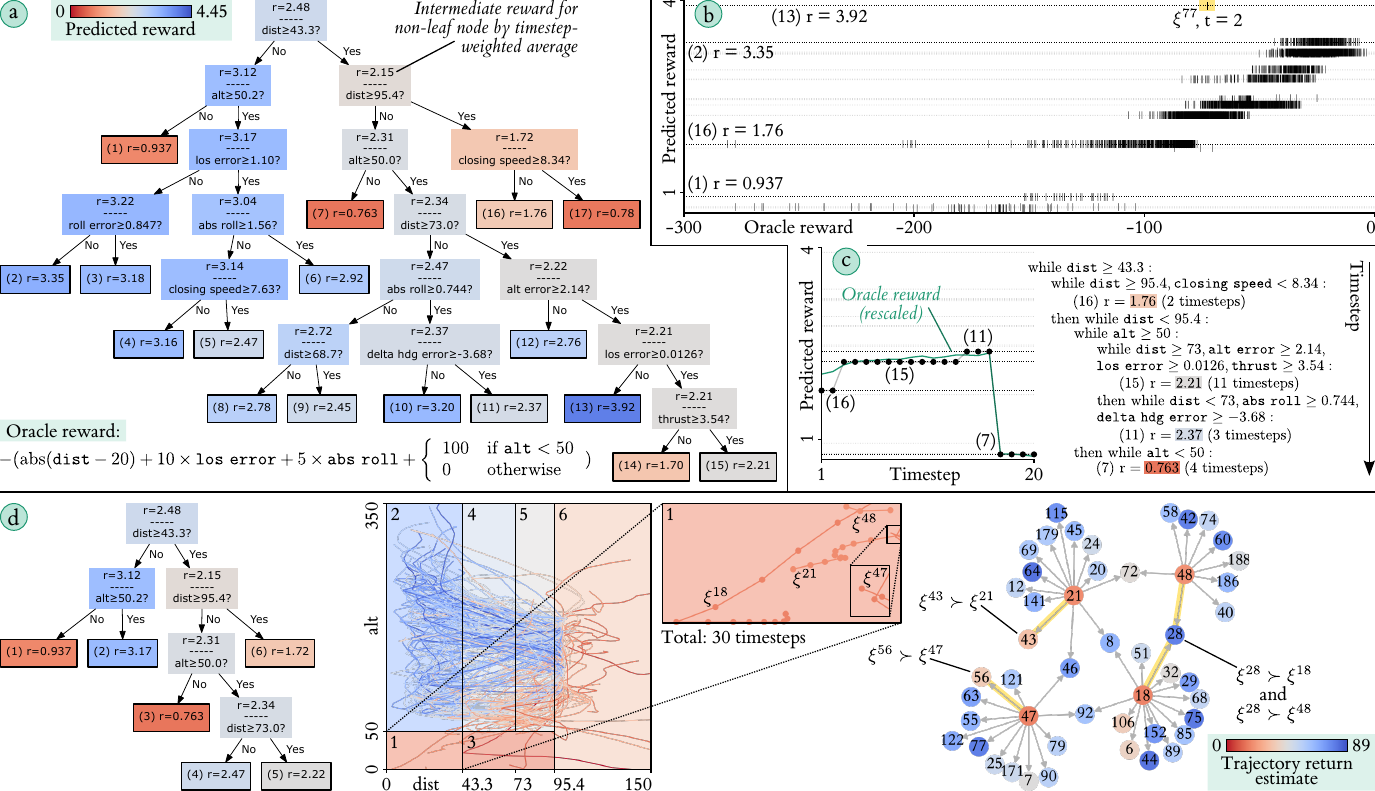}
\caption{Analysis of a reward tree learnt for the Chase task.}
\label{fig:tree_analysis}
\end{figure}

We have shown that reward learning with $\ell_{\text{0-1}}$-based trees can be competitive with NNs, but not quite as performant overall. We now turn to a concrete advantage which may tip practical
% calculations
trade-offs
in its favour: the ability to interpret the learnt model, and analyse how its predictions relate to the underlying preference graph. In Figure \ref{fig:tree_analysis} we opt for depth over breadth, and focus on the single best tree by ORR on the Chase task. The figure is divided into sections (\textbf{a} -- \textbf{d}):

(\textbf{a})\quad This reward tree has $17$ leaves.
%defined by a hierarchy of rules over the feature space.
The oracle reward, printed below, uses four features, all of which are used in the tree in ways that are broadly aligned (e.g. lower \texttt{los} \texttt{error} leads to leaves with higher reward). The model has learnt the crucial threshold $\texttt{alt}<50$, correctly assigning a low reward when it is crossed. This explains why we observe rare violations of the altitude threshold in Figure \ref{fig:trajectory_visualisation}. However, it has not learnt the ideal distance to RJ, $\texttt{dist}=20$, with $43.3$ being the lowest value used in a rule. This could be because the underlying preference graph lacks sufficient preferences to make this distinction; adopting an active querying scheme may help to discover such subtleties efficiently. Other features besides those used by the oracle are present in the tree, indicating some causal confusion \citep{tien2022study}. This may not necessarily harm agent performance, as it could provide beneficial shaping (e.g. penalising positive \texttt{closing} \texttt{speed}, which indicates increasing distance to RJ). That may indeed be the case for this model since ORR is actually negative.

(\textbf{b})\quad We plot the tree's predicted reward against the oracle reward for all timesteps in the online trajectories (correlation $=0.903$). The predictions for each leaf lie along a horizontal line. Most leaves, including $1$ and $2$, are well-aligned on this data because their oracle reward distributions are tightly concentrated around low/high averages respectively (note that the absolute scale is irrelevant here). Leaf $16$ has a wider oracle reward distribution, with several negative outliers. An optimal tree would likely split this leaf further, perhaps using the $\texttt{alt}<50$ threshold.
%In general, alignment appears to be better for timesteps with higher oracle reward, likely because more examples are present in the online data. This may be good; a limited capacity tree better to distinguish good and very good behaviour, as opposed to poor and very poor.
The one anomaly is leaf $13$, which contains just a single timestep from $\xi^{77}$. This trajectory is the eighth best in the dataset by oracle return, but this leaf assigns that credit to a state that seemingly does not merit it, as the distance to RJ is so high ($\texttt{dist}>73$). This may be an example of suboptimal reward learning, but the fact that its origin can be pinpointed so precisely is a testament to the value of interpretability.

(\textbf{c})\quad We leverage the tree structure to produce a human-readable explanation of reward predictions for a single trajectory, which may be of value to an end user (e.g. a trainee pilot seeking to understand the strengths and weaknesses of their own performance). We consider $\xi^{191}$, a rare case that violates the altitude threshold. The time series of reward shows that the $20$ timesteps are spent in leaves $16$, $15$, $11$ and $7$. Rescaled oracle rewards are overlaid in teal, and show that the model's predictions are well-aligned. To the right, we translate this visualisation into a textual form, similar to a nested program. Read top-to-bottom, the text indicates which rules of the tree are active at each timestep, and the effect this has on the predicted reward. This trajectory starts fairly positively, with reward gradually increasing over the first $16$ timesteps as \texttt{dist} is reduced to between $43.3$ and $73$, but then falls dramatically when the $\texttt{alt}<50$ threshold is crossed. We are unaware of any method that could extract such a compact explanation of sequential predictions from an NN.

(\textbf{d})\quad We isolate a subtree, starting at the root node, that splits only on \texttt{dist} and \texttt{alt}. We give a spatial representation of the subtree, and how it is populated by the $200$ online trajectories, using a 2D partition plot analogous to those in Figure \ref{fig:problem_and_solution}. Zooming into leaf $1$, which covers cases where the altitude threshold is violated, we see that it contains a total of $30$ timesteps across four trajectories. By Equation \ref{eq:reward_prediction}, the low reward for this leaf results from a weighted average of the return estimates for these four trajectories, which in turn (by Equation \ref{eq:proxy_loss_trajectory_level}) are derived from the preference graph. We can use this inverse reasoning to explain \textit{why} this leaf has much lower reward than its sibling (leaf $2$ of the subtree). A proximal explanation comes by filtering the graph for preferences that specifically compare trajectories that visit those two leaves. $49$ such preferences exist, and in all cases, the oracle prefers the trajectory that does not visit leaf $1$. Some of these preferences may be more practically salient than others. For example, we might highlight trajectories that feature more than once (e.g. $\xi^{28}$ is preferred to both $\xi^{18}$ and $\xi^{48}$), or cases where trajectories with low overall return estimates are nonetheless preferred to those in leaf $1$ (e.g. $\xi^{43}\succ \xi^{21}$ and $\xi^{56}\succ \xi^{47}$).
This ability to trace a reward tree's predictions back to the individual preferences that influence them could be valuable for verification and active learning.
% We believe that much more could be done to extend and apply this basic framework for the traceable explanation of preference-based reward.

\vspace{-0.4cm}
\section{Conclusions and Future Work}
\label{sec:conclusion}
\vspace{0.2cm}

Fast jet handling is an excellent case study of both the value and difficulty of distilling tacit human expertise into software systems.
In this work, we proposed a preference-based reward learning framework, which yields both a quantitative model of human preferences with a readable tree structure, and an artificial demonstrator agent capable of executing high-quality flight trajectories with respect to that model.
The intrinsic interpretability of the reward tree model enables improved insight into the structure of expert preferences, and verification of agent behaviour, compared with standard NN-based approaches.
Through oracle experiments on several aircraft handling tasks, we showed that reward trees with around $20$ leaves can achieve quantitative and qualitative performance close to that of NNs, with a more direct split criterion bringing consistent benefits. We found that the NN-tree gap reduces as the true reward becomes more nonlinear, and remains stable or reduces further in the presence of limited or corrupted data.

Although our experiments used synthetic oracle data to enable scalable quantitative evaluation, future work should include studies using the preferences of real human experts in the aircraft handling domain.
It should also be noted that any realistic preference elicitation scenario would likely involve multiple experts with differing knowledge and expertise. A natural extension of our approach, which we see as valuable future work, is to learn individual reward functions for each expert, then leverage the intrinsic interpretability to identify biases, inconsistencies and trade-offs. This suggests a further application of reward tree learning: providing a basis for evaluating and training the
experts themselves.

\newpage 

\section*{Appendix}
\appendix

\begin{table}[h!]
\centering
\small
\caption{List of features used by oracles and reward learning models.
Apart from those containing ``\texttt{delta}" or ``\texttt{rate}", all features are computed over the successor state for each transition, $s_{t+1}$.}
\label{tab:features}
\begin{tabular}{|c|c|}
\texttt{dist}                 & Euclidean distance between EJ and RJ \\
\texttt{closing}\ \texttt{speed}       & Closing speed between EJ and RJ (negative = moving closer) \\
\texttt{alt}                  & Altitude of EJ \\
\texttt{alt}\ \texttt{error}           & Difference in altitude between EJ and RJ (negative = EJ is lower) \\
\texttt{delta}\ \texttt{alt}\ \texttt{error}    & Change in \texttt{alt}\ \texttt{error} between $s_t$ and $s_{t+1}$ \\
\texttt{dist}\ \texttt{hor}             & Euclidean distance between EJ and RJ in horizontal plane \\
\texttt{delta}\ \texttt{dist}\ \texttt{hor}      &  Change in \texttt{dist}\ \texttt{hor} between $s_t$ and $s_{t+1}$ (negative = moving closer) \\
\texttt{pitch}\ \texttt{error}         & Absolute difference in pitch angle between EJ and RJ \\
\texttt{delta}\ \texttt{pitch}\ \texttt{error}  & Change in \texttt{pitch}\ \texttt{error} between $s_t$ and $s_{t+1}$ \\
\texttt{abs}\ \texttt{roll}            & Absolute roll angle of EJ \\
\texttt{roll}\ \texttt{error}          & Absolute difference in roll angle between EJ and RJ \\
\texttt{delta}\ \texttt{roll}\ \texttt{error}   & Change in \texttt{roll}\ \texttt{error} between $s_t$ and $s_{t+1}$ \\
\texttt{hdg}\ \texttt{error}           & Absolute difference in heading angle between EJ and RJ \\
\texttt{delta}\ \texttt{hdg}\ \texttt{error}    & Change in \texttt{hdg}\ \texttt{error} between $s_t$ and $s_{t+1}$ \\
\texttt{fwd}\ \texttt{error}           & Angle between 3D vectors indicating forward axes of EJ and RJ \\
\texttt{delta}\ \texttt{fwd}\ \texttt{error}    &  Change in \texttt{fwd}\ \texttt{error} between $s_t$ and $s_{t+1}$ \\
\texttt{up}\ \texttt{error}            & Angle between 3D vectors indicating upward axes of EJ and RJ \\
\texttt{delta}\ \texttt{up}\ \texttt{error}     & Change in \texttt{up}\ \texttt{error} between $s_t$ and $s_{t+1}$ \\
\texttt{right}\ \texttt{error}         & Angle between 3D vectors indicating rightward axes of EJ and RJ \\
\texttt{delta}\ \texttt{right}\ \texttt{error}  & Change in \texttt{right}\ \texttt{error} between $s_t$ and $s_{t+1}$ \\
\texttt{los}\ \texttt{error}           & Angle between forward axis of EJ and vector from EJ to RJ \\ & (measures whether RJ is in EJ's line of sight) \\
\texttt{delta}\ \texttt{los}\ \texttt{error}    & Change in \texttt{los}\ \texttt{error} between $s_t$ and $s_{t+1}$ \\
\texttt{abs}\ \texttt{lr}\ \texttt{offset}      & Magnitude of projection of vector from EJ to RJ onto RJ's rightward axis \\ & (measures left-right offset between the two aircraft in RJ's reference frame) \\
\texttt{speed}                & Airspeed of EJ \\
\texttt{g}\ \texttt{force}             & Instantaneous g-force experienced by EJ \\
\texttt{pitch}\ \texttt{rate}          & Absolute change of EJ pitch between $s_t$ and $s_{t+1}$ \\
\texttt{roll}\ \texttt{rate}           & Absolute change of EJ roll between $s_t$ and $s_{t+1}$ \\
\texttt{yaw}\ \texttt{rate}            & Absolute change of EJ yaw between $s_t$ and $s_{t+1}$ \\
\texttt{thrust}               & Instantaneous thrust output by EJ engines \\
\texttt{delta}\ \texttt{thrust}        & Absolute change in \texttt{thrust} between $s_t$ and $s_{t+1}$
\end{tabular}
\end{table}

\section*{Acknowledgments}
\vspace{0.2cm}

This work was supported by a Thales/EPSRC Industrial Case Award in autonomous systems.
The aircraft simulator used in experiments was based on an initial implementation by Ian Henderson at Thales UK, and experimental tasks were developed in consultation with both Ian Henderson and Rachel Craddock, also at Thales UK.

\newpage

\bibliography{bibliography}

\begin{thebibliography}{72}
\newcommand{\enquote}[1]{``#1''}
\providecommand{\natexlab}[1]{#1}
\providecommand{\url}[1]{\texttt{#1}}
\providecommand{\urlprefix}{URL }
\expandafter\ifx\csname urlstyle\endcsname\relax
  \providecommand{\doi}[1]{\discretionary{}{}{}https://doi.org/#1}\else
  \providecommand{\doi}[1]{\discretionary{}{}{}\urlstyle{rm}\url{https://doi.org/#1}}\fi

\bibitem[{Sternberg and Horvath(1999)}]{sternberg1999tacit}
Sternberg, R.~J., and Horvath, J.~A., \emph{Tacit knowledge in professional
  practice: Researcher and practitioner perspectives}, Psychology Press, 1999.

\bibitem[{Rudin(2019)}]{rudin2019stop}
Rudin, C., \enquote{Stop explaining black box machine learning models for high
  stakes decisions and use interpretable models instead,} \emph{Nature Machine
  Intelligence}, Vol.~1, No.~5, 2019, pp. 206--215.

\bibitem[{Brunton et~al.(2021)Brunton, Nathan~Kutz, Manohar, Aravkin,
  Morgansen, Klemisch, Goebel, Buttrick, Poskin, Blom-Schieber
  et~al.}]{brunton2021data}
Brunton, S.~L., Nathan~Kutz, J., Manohar, K., Aravkin, A.~Y., Morgansen, K.,
  Klemisch, J., Goebel, N., Buttrick, J., Poskin, J., Blom-Schieber, A.~W.,
  et~al., \enquote{Data-driven aerospace engineering: reframing the industry
  with machine learning,} \emph{AIAA Journal}, Vol.~59, No.~8, 2021, pp.
  2820--2847.

\bibitem[{Bewley and Lecue(2022)}]{bewley2022interpretable}
Bewley, T., and Lecue, F., \enquote{Interpretable Preference-based
  Reinforcement Learning with Tree-Structured Reward Functions,}
  \emph{Proceedings of the 21st International Conference on Autonomous Agents
  and Multiagent Systems}, 2022, pp. 118--126.

\bibitem[{Christiano et~al.(2017)Christiano, Leike, Brown, Martic, Legg, and
  Amodei}]{christiano2017deep}
Christiano, P.~F., Leike, J., Brown, T., Martic, M., Legg, S., and Amodei, D.,
  \enquote{Deep reinforcement learning from human preferences,} \emph{Advances
  in Neural Information Processing Systems}, Vol.~30, 2017.

\bibitem[{Lee et~al.(2021{\natexlab{a}})Lee, Smith, and Abbeel}]{lee2021pebble}
Lee, K., Smith, L.~M., and Abbeel, P., \enquote{PEBBLE: Feedback-Efficient
  Interactive Reinforcement Learning via Relabeling Experience and Unsupervised
  Pre-training,} \emph{International Conference on Machine Learning}, PMLR,
  2021{\natexlab{a}}, pp. 6152--6163.

\bibitem[{Griffith et~al.(2013)Griffith, Subramanian, Scholz, Isbell, and
  Thomaz}]{griffith2013policy}
Griffith, S., Subramanian, K., Scholz, J., Isbell, C.~L., and Thomaz, A.~L.,
  \enquote{Policy shaping: Integrating human feedback with reinforcement
  learning,} \emph{Advances in neural information processing systems}, Vol.~26,
  2013.

\bibitem[{Reddy et~al.(2020)Reddy, Dragan, Levine, Legg, and
  Leike}]{reddy2020learning}
Reddy, S., Dragan, A., Levine, S., Legg, S., and Leike, J., \enquote{Learning
  human objectives by evaluating hypothetical behavior,} \emph{International
  Conference on Machine Learning}, PMLR, 2020, pp. 8020--8029.

\bibitem[{Lindner et~al.(2021)Lindner, Turchetta, Tschiatschek, Ciosek, and
  Krause}]{lindner2021information}
Lindner, D., Turchetta, M., Tschiatschek, S., Ciosek, K., and Krause, A.,
  \enquote{Information Directed Reward Learning for Reinforcement Learning,}
  \emph{Advances in Neural Information Processing Systems}, Vol.~34, 2021, pp.
  3850--3862.

\bibitem[{Lee et~al.(2021{\natexlab{b}})Lee, Smith, Dragan, and
  Abbeel}]{lee2021bpref}
Lee, K., Smith, L., Dragan, A., and Abbeel, P., \enquote{B-Pref: Benchmarking
  Preference-Based Reinforcement Learning,} \emph{Advances in Neural
  Information Processing Systems}, Vol.~35, 2021{\natexlab{b}}.

\bibitem[{Sutton and Barto(2018)}]{sutton2018reinforcement}
Sutton, R.~S., and Barto, A.~G., \emph{Reinforcement learning: An
  introduction}, MIT press, 2018.

\bibitem[{Azar et~al.(2021)Azar, Koubaa, Ali~Mohamed, Ibrahim, Ibrahim, Kazim,
  Ammar, Benjdira, Khamis, Hameed et~al.}]{azar2021drone}
Azar, A.~T., Koubaa, A., Ali~Mohamed, N., Ibrahim, H.~A., Ibrahim, Z.~F.,
  Kazim, M., Ammar, A., Benjdira, B., Khamis, A.~M., Hameed, I.~A., et~al.,
  \enquote{Drone deep reinforcement learning: A review,} \emph{Electronics},
  Vol.~10, No.~9, 2021, p. 999.

\bibitem[{Liu et~al.(2022)Liu, Kiumarsi, Kartal, Koru, Modares, and
  Lewis}]{liu2022reinforcement}
Liu, H., Kiumarsi, B., Kartal, Y., Koru, A.~T., Modares, H., and Lewis, F.~L.,
  \enquote{Reinforcement learning applications in unmanned vehicle control: A
  comprehensive overview,} \emph{Unmanned Systems}, 2022, pp. 1--10.

\bibitem[{Razzaghi et~al.(2022)Razzaghi, Tabrizian, Guo, Chen, Taye, Thompson,
  Bregeon, Baheri, and Wei}]{razzaghi2022survey}
Razzaghi, P., Tabrizian, A., Guo, W., Chen, S., Taye, A., Thompson, E.,
  Bregeon, A., Baheri, A., and Wei, P., \enquote{A Survey on Reinforcement
  Learning in Aviation Applications,} \emph{arXiv preprint arXiv:2211.02147},
  2022.

\bibitem[{Tang and Lai(2020)}]{tang2020deep}
Tang, C., and Lai, Y.-C., \enquote{Deep reinforcement learning automatic
  landing control of fixed-wing aircraft using deep deterministic policy
  gradient,} \emph{2020 International Conference on Unmanned Aircraft Systems
  (ICUAS)}, IEEE, 2020, pp. 1--9.

\bibitem[{Clarke and Hwang(2020)}]{clarke2020deep}
Clarke, S.~G., and Hwang, I., \enquote{Deep reinforcement learning control for
  aerobatic maneuvering of agile fixed-wing aircraft,} \emph{AIAA Scitech 2020
  Forum}, 2020, p. 0136.

\bibitem[{Morales and Sammut(2004)}]{morales2004learning}
Morales, E.~F., and Sammut, C., \enquote{Learning to fly by combining
  reinforcement learning with behavioural cloning,} \emph{Proceedings of the
  twenty-first international conference on Machine learning}, 2004, p.~76.

\bibitem[{Cao et~al.(2022)Cao, Wang, Zhang, Yu, and
  Shen}]{cao2022demonstration}
Cao, S., Wang, X., Zhang, R., Yu, H., and Shen, L., \enquote{From Demonstration
  to Flight: Realization of Autonomous Aerobatic Maneuvers for Fast, Miniature
  Fixed-Wing UAVs,} \emph{IEEE Robotics and Automation Letters}, Vol.~7, No.~2,
  2022, pp. 5771--5778.

\bibitem[{Yildiz et~al.(2014)Yildiz, Agogino, and Brat}]{yildiz2014predicting}
Yildiz, Y., Agogino, A., and Brat, G., \enquote{Predicting pilot behavior in
  medium-scale scenarios using game theory and reinforcement learning,}
  \emph{Journal of Guidance, Control, and Dynamics}, Vol.~37, No.~4, 2014, pp.
  1335--1343.

\bibitem[{Vemuru et~al.(2019)Vemuru, Harbour, and
  Clark}]{vemuru2019reinforcement}
Vemuru, K.~V., Harbour, S.~D., and Clark, J.~D., \enquote{Reinforcement
  Learning in Aviation, Either Unmanned or Manned, with an Injection of AI,}
  \emph{20th International Symposium on Aviation Psychology}, 2019, p. 492.

\bibitem[{van Oijen et~al.(2017)van Oijen, Poppinga, Brouwer, Aliko, and
  Roessingh}]{van2017towards}
van Oijen, J., Poppinga, G., Brouwer, O., Aliko, A., and Roessingh, J.~J.,
  \enquote{Towards modeling the learning process of aviators using deep
  reinforcement learning,} \emph{2017 IEEE International Conference on Systems,
  Man, and Cybernetics (SMC)}, IEEE, 2017, pp. 3439--3444.

\bibitem[{Russell(2019)}]{russell2019human}
Russell, S., \emph{Human compatible: Artificial intelligence and the problem of
  control}, Penguin, 2019.

\bibitem[{Leike et~al.(2018)Leike, Krueger, Everitt, Martic, Maini, and
  Legg}]{leike2018scalable}
Leike, J., Krueger, D., Everitt, T., Martic, M., Maini, V., and Legg, S.,
  \enquote{Scalable agent alignment via reward modeling: a research direction,}
  \emph{arXiv preprint arXiv:1811.07871}, 2018.

\bibitem[{Ng et~al.(2000)Ng, Russell et~al.}]{ng2000algorithms}
Ng, A.~Y., Russell, S., et~al., \enquote{Algorithms for inverse reinforcement
  learning.} \emph{Icml}, Vol.~1, 2000, p.~2.

\bibitem[{Knox and Stone(2008)}]{knox2008tamer}
Knox, W.~B., and Stone, P., \enquote{Tamer: Training an agent manually via
  evaluative reinforcement,} \emph{2008 7th IEEE international conference on
  development and learning}, IEEE, 2008, pp. 292--297.

\bibitem[{Bajcsy et~al.(2017)Bajcsy, Losey, O’Malley, and
  Dragan}]{bajcsy2017learning}
Bajcsy, A., Losey, D.~P., O’Malley, M.~K., and Dragan, A.~D.,
  \enquote{Learning robot objectives from physical human interaction,}
  \emph{Conference on Robot Learning}, PMLR, 2017, pp. 217--226.

\bibitem[{Wirth et~al.(2016)Wirth, F{\"u}rnkranz, and Neumann}]{wirth2016model}
Wirth, C., F{\"u}rnkranz, J., and Neumann, G., \enquote{Model-free
  preference-based reinforcement learning,} \emph{Thirtieth AAAI Conference on
  Artificial Intelligence}, 2016.

\bibitem[{Sadigh et~al.(2017)Sadigh, Dragan, Sastry, and
  Seshia}]{sadigh2017active}
Sadigh, D., Dragan, A.~D., Sastry, S., and Seshia, S.~A., \enquote{Active
  preference-based learning of reward functions,} \emph{Proceedings of
  Robotics: Science and Systems (RSS)}, 2017.

\bibitem[{Cao et~al.(2021)Cao, Wong, and Lin}]{cao2021weak}
Cao, Z., Wong, K., and Lin, C.-T., \enquote{Weak human preference supervision
  for deep reinforcement learning,} \emph{IEEE Transactions on Neural Networks
  and Learning Systems}, Vol.~32, No.~12, 2021, pp. 5369--5378.

\bibitem[{Kendall(1975)}]{kendall1975kendall}
Kendall, M., \enquote{{Rank Correlation Methods; Griffin, C., Ed},} , 1975.

\bibitem[{Wilde et~al.(2020)Wilde, Blidaru, Smith, and
  Kuli{\'c}}]{wilde2020improving}
Wilde, N., Blidaru, A., Smith, S.~L., and Kuli{\'c}, D., \enquote{Improving
  user specifications for robot behavior through active preference learning:
  Framework and evaluation,} \emph{The International Journal of Robotics
  Research}, Vol.~39, No.~6, 2020, pp. 651--667.

\bibitem[{Guo et~al.(2018)Guo, Tian, Kalpathy-Cramer, Ostmo, Campbell, Chiang,
  Erdogmus, Dy, and Ioannidis}]{guo2018experimental}
Guo, Y., Tian, P., Kalpathy-Cramer, J., Ostmo, S., Campbell, J.~P., Chiang,
  M.~F., Erdogmus, D., Dy, J.~G., and Ioannidis, S., \enquote{Experimental
  Design under the Bradley-Terry Model.} \emph{IJCAI}, 2018, pp. 2198--2204.

\bibitem[{Ibarz et~al.(2018)Ibarz, Leike, Pohlen, Irving, Legg, and
  Amodei}]{ibarz2018reward}
Ibarz, B., Leike, J., Pohlen, T., Irving, G., Legg, S., and Amodei, D.,
  \enquote{Reward learning from human preferences and demonstrations in Atari,}
  \emph{Advances in Neural Information Processing Systems}, Vol.~31, 2018.

\bibitem[{B{\i}y{\i}k et~al.(2022)B{\i}y{\i}k, Losey, Palan, Landolfi,
  Shevchuk, and Sadigh}]{biyik2022learning}
B{\i}y{\i}k, E., Losey, D.~P., Palan, M., Landolfi, N.~C., Shevchuk, G., and
  Sadigh, D., \enquote{Learning reward functions from diverse sources of human
  feedback: Optimally integrating demonstrations and preferences,} \emph{The
  International Journal of Robotics Research}, Vol.~41, No.~1, 2022, pp.
  45--67.

\bibitem[{Puiutta and Veith(2020)}]{puiutta2020explainable}
Puiutta, E., and Veith, E., \enquote{Explainable reinforcement learning: A
  survey,} \emph{International cross-domain conference for machine learning and
  knowledge extraction}, Springer, 2020, pp. 77--95.

\bibitem[{Heuillet et~al.(2021)Heuillet, Couthouis, and
  D{\'\i}az-Rodr{\'\i}guez}]{heuillet2021explainability}
Heuillet, A., Couthouis, F., and D{\'\i}az-Rodr{\'\i}guez, N.,
  \enquote{Explainability in deep reinforcement learning,}
  \emph{Knowledge-Based Systems}, Vol. 214, 2021, p. 106685.

\bibitem[{Zhu et~al.(2018)Zhu, Huang, and Zhang}]{zhu2018object}
Zhu, G., Huang, Z., and Zhang, C., \enquote{Object-oriented dynamics
  predictor,} \emph{Advances in Neural Information Processing Systems},
  Vol.~31, 2018.

\bibitem[{Verma et~al.(2018)Verma, Murali, Singh, Kohli, and
  Chaudhuri}]{verma2018programmatically}
Verma, A., Murali, V., Singh, R., Kohli, P., and Chaudhuri, S.,
  \enquote{Programmatically interpretable reinforcement learning,}
  \emph{International Conference on Machine Learning}, PMLR, 2018, pp.
  5045--5054.

\bibitem[{Zahavy et~al.(2016)Zahavy, Ben-Zrihem, and
  Mannor}]{zahavy2016graying}
Zahavy, T., Ben-Zrihem, N., and Mannor, S., \enquote{Graying the black box:
  Understanding dqns,} \emph{International conference on machine learning},
  PMLR, 2016, pp. 1899--1908.

\bibitem[{Huber et~al.(2019)Huber, Schiller, and
  Andr{\'e}}]{huber2019enhancing}
Huber, T., Schiller, D., and Andr{\'e}, E., \enquote{Enhancing explainability
  of deep reinforcement learning through selective layer-wise relevance
  propagation,} \emph{Joint German/Austrian Conference on Artificial
  Intelligence (K{\"u}nstliche Intelligenz)}, Springer, 2019, pp. 188--202.

\bibitem[{van~der Waa et~al.(2018)van~der Waa, van Diggelen, Bosch, and
  Neerincx}]{van2018contrastive}
van~der Waa, J., van Diggelen, J., Bosch, K. v.~d., and Neerincx, M.,
  \enquote{Contrastive explanations for reinforcement learning in terms of
  expected consequences,} \emph{IJCAI/ECAI Workshop on Explainable Artificial
  Intelligence}, 2018.

\bibitem[{Amir and Amir(2018)}]{amir2018highlights}
Amir, D., and Amir, O., \enquote{Highlights: Summarizing agent behavior to
  people,} \emph{Proceedings of the 17th International Conference on Autonomous
  Agents and MultiAgent Systems}, 2018.

\bibitem[{Huang et~al.(2018)Huang, Bhatia, Abbeel, and
  Dragan}]{huang2018establishing}
Huang, S.~H., Bhatia, K., Abbeel, P., and Dragan, A.~D., \enquote{Establishing
  appropriate trust via critical states,} \emph{2018 IEEE/RSJ International
  Conference on Intelligent Robots and Systems (IROS)}, IEEE, 2018, pp.
  3929--3936.

\bibitem[{Dao et~al.(2018)Dao, Mishra, and Lee}]{dao2018deep}
Dao, G., Mishra, I., and Lee, M., \enquote{Deep reinforcement learning monitor
  for snapshot recording,} \emph{2018 17th IEEE International Conference on
  Machine Learning and Applications (ICMLA)}, IEEE, 2018, pp. 591--598.

\bibitem[{Bewley et~al.(2022)Bewley, Lawry, and
  Richards}]{bewley2022summarising}
Bewley, T., Lawry, J., and Richards, A., \enquote{Summarising and Comparing
  Agent Dynamics with Contrastive Spatiotemporal Abstraction,} \emph{IJCAI/ECAI
  Workshop on Explainable Artificial Intelligence}, 2022.

\bibitem[{Glanois et~al.(2021)Glanois, Weng, Zimmer, Li, Yang, Hao, and
  Liu}]{glanois2021survey}
Glanois, C., Weng, P., Zimmer, M., Li, D., Yang, T., Hao, J., and Liu, W.,
  \enquote{A Survey on Interpretable Reinforcement Learning,} \emph{arXiv
  preprint arXiv:2112.13112}, 2021.

\bibitem[{Juozapaitis et~al.(2019)Juozapaitis, Koul, Fern, Erwig, and
  Doshi-Velez}]{juozapaitis2019explainable}
Juozapaitis, Z., Koul, A., Fern, A., Erwig, M., and Doshi-Velez, F.,
  \enquote{Explainable reinforcement learning via reward decomposition,}
  \emph{IJCAI/ECAI Workshop on Explainable Artificial Intelligence}, 2019.

\bibitem[{Devidze et~al.(2021)Devidze, Radanovic, Kamalaruban, and
  Singla}]{devidze2021explicable}
Devidze, R., Radanovic, G., Kamalaruban, P., and Singla, A.,
  \enquote{Explicable reward design for reinforcement learning agents,}
  \emph{Advances in Neural Information Processing Systems}, Vol.~34, 2021, pp.
  20118--20131.

\bibitem[{Russell and Santos(2019)}]{russell2019explaining}
Russell, J., and Santos, E., \enquote{{Explaining reward functions in Markov
  decision processes},} \emph{The Thirty-Second International Flairs
  Conference}, 2019.

\bibitem[{Michaud et~al.(2020)Michaud, Gleave, and
  Russell}]{michaud2020understanding}
Michaud, E.~J., Gleave, A., and Russell, S., \enquote{Understanding learned
  reward functions,} \emph{arXiv preprint arXiv:2012.05862}, 2020.

\bibitem[{Jenner and Gleave(2022)}]{jenner2022preprocessing}
Jenner, E., and Gleave, A., \enquote{Preprocessing Reward Functions for
  Interpretability,} \emph{arXiv preprint arXiv:2203.13553}, 2022.

\bibitem[{Chapman and Kaelbling(1991)}]{chapman1991input}
Chapman, D., and Kaelbling, L.~P., \enquote{Input Generalization in Delayed
  Reinforcement Learning: An Algorithm and Performance Comparisons.}
  \emph{Ijcai}, Vol.~91, 1991, pp. 726--731.

\bibitem[{D{\v{z}}eroski et~al.(1998)D{\v{z}}eroski, Raedt, and
  Blockeel}]{dvzeroski1998relational}
D{\v{z}}eroski, S., Raedt, L.~D., and Blockeel, H., \enquote{Relational
  reinforcement learning,} \emph{International Conference on Inductive Logic
  Programming}, Springer, 1998, pp. 11--22.

\bibitem[{Pyeatt(2003)}]{pyeatt2003reinforcement}
Pyeatt, L.~D., \enquote{Reinforcement learning with decision trees.} \emph{21
  st IASTED International Multi-Conference on Applied Informatics}, 2003, pp.
  26--31.

\bibitem[{Silva et~al.(2020)Silva, Gombolay, Killian, Jimenez, and
  Son}]{pmlr-v108-silva20a}
Silva, A., Gombolay, M., Killian, T., Jimenez, I., and Son, S.-H.,
  \enquote{Optimization Methods for Interpretable Differentiable Decision Trees
  Applied to Reinforcement Learning,} \emph{Proceedings of the Twenty Third
  International Conference on Artificial Intelligence and Statistics}, Vol.
  108, PMLR, 2020, pp. 1855--1865.

\bibitem[{Liu et~al.(2018)Liu, Schulte, Zhu, and Li}]{liu2018toward}
Liu, G., Schulte, O., Zhu, W., and Li, Q., \enquote{Toward interpretable deep
  reinforcement learning with linear model u-trees,} \emph{Joint European
  Conference on Machine Learning and Knowledge Discovery in Databases},
  Springer, 2018, pp. 414--429.

\bibitem[{Jiang et~al.(2019)Jiang, Hwang, and Lin}]{jiang2019experience}
Jiang, W.-C., Hwang, K.-S., and Lin, J.-L., \enquote{An experience replay
  method based on tree structure for reinforcement learning,} \emph{IEEE
  Transactions on Emerging Topics in Computing}, Vol.~9, No.~2, 2019, pp.
  972--982.

\bibitem[{Bastani et~al.(2018)Bastani, Pu, and
  Solar-Lezama}]{bastani2018verifiable}
Bastani, O., Pu, Y., and Solar-Lezama, A., \enquote{Verifiable Reinforcement
  Learning via Policy Extraction,} \emph{Advances in Neural Information
  Processing Systems}, Vol.~31, 2018.

\bibitem[{Cobo et~al.(2012)Cobo, Isbell~Jr, and Thomaz}]{cobo2012automatic}
Cobo, L.~C., Isbell~Jr, C.~L., and Thomaz, A.~L., \enquote{Automatic task
  decomposition and state abstraction from demonstration,} Georgia Institute of
  Technology, 2012.

\bibitem[{Tambwekar et~al.(2021)Tambwekar, Silva, Gopalan, and
  Gombolay}]{tambwekar2021specifying}
Tambwekar, P., Silva, A., Gopalan, N., and Gombolay, M., \enquote{Specifying
  and Interpreting Reinforcement Learning Policies through Simulatable Machine
  Learning,} \emph{arXiv preprint arXiv:2101.07140}, 2021.

\bibitem[{Sheikh et~al.(2022)Sheikh, Khadka, Miret, Majumdar, and
  Phielipp}]{sheikh2022learning}
Sheikh, H.~U., Khadka, S., Miret, S., Majumdar, S., and Phielipp, M.,
  \enquote{Learning intrinsic symbolic rewards in reinforcement learning,}
  \emph{2022 International Joint Conference on Neural Networks (IJCNN)}, IEEE,
  2022, pp. 1--8.

\bibitem[{Bradley and Terry(1952)}]{bradley1952rank}
Bradley, R.~A., and Terry, M.~E., \enquote{Rank analysis of incomplete block
  designs: I. The method of paired comparisons,} \emph{Biometrika}, Vol.~39,
  No. 3/4, 1952, pp. 324--345.

\bibitem[{Su{\'a}rez and Lutsko(1999)}]{suarez1999globally}
Su{\'a}rez, A., and Lutsko, J.~F., \enquote{Globally optimal fuzzy decision
  trees for classification and regression,} \emph{IEEE Transactions on Pattern
  Analysis and Machine Intelligence}, Vol.~21, No.~12, 1999, pp. 1297--1311.

\bibitem[{Gulliksen(1956)}]{gulliksen1956least}
Gulliksen, H., \enquote{A least squares solution for paired comparisons with
  incomplete data,} \emph{Psychometrika}, Vol.~21, No.~2, 1956, pp. 125--134.

\bibitem[{Kingma and Ba(2014)}]{kingma2014adam}
Kingma, D.~P., and Ba, J., \enquote{Adam: A method for stochastic
  optimization,} \emph{arXiv preprint arXiv:1412.6980}, 2014.

\bibitem[{Breiman et~al.(2017)Breiman, Friedman, Olshen, and
  Stone}]{breiman2017classification}
Breiman, L., Friedman, J.~H., Olshen, R.~A., and Stone, C.~J.,
  \emph{Classification and regression trees}, Routledge, 2017.

\bibitem[{Tien et~al.(2022)Tien, He, Erickson, Dragan, and
  Brown}]{tien2022study}
Tien, J., He, J. Z.-Y., Erickson, Z., Dragan, A.~D., and Brown, D., \enquote{A
  Study of Causal Confusion in Preference-Based Reward Learning,} \emph{arXiv
  preprint arXiv:2204.06601}, 2022.

\bibitem[{Haarnoja et~al.(2018)Haarnoja, Zhou, Abbeel, and
  Levine}]{haarnoja2018soft}
Haarnoja, T., Zhou, A., Abbeel, P., and Levine, S., \enquote{Soft actor-critic:
  Off-policy maximum entropy deep reinforcement learning with a stochastic
  actor,} \emph{International Conference on Machine Learning}, PMLR, 2018, pp.
  1861--1870.

\bibitem[{Rahtz et~al.(2022)Rahtz, Varma, Kumar, Kenton, Legg, and
  Leike}]{rahtz2022safe}
Rahtz, M., Varma, V., Kumar, R., Kenton, Z., Legg, S., and Leike, J.,
  \enquote{Safe Deep RL in 3D Environments using Human Feedback,} \emph{arXiv
  preprint arXiv:2201.08102}, 2022.

\bibitem[{Chua et~al.(2018)Chua, Calandra, McAllister, and
  Levine}]{chua2018deep}
Chua, K., Calandra, R., McAllister, R., and Levine, S., \enquote{Deep
  reinforcement learning in a handful of trials using probabilistic dynamics
  models,} \emph{Advances in Neural Information Processing Systems}, Vol.~31,
  2018.

\bibitem[{Gleave et~al.(2021)Gleave, Dennis, Legg, Russell, and
  Leike}]{gleave2021quantifying}
Gleave, A., Dennis, M.~D., Legg, S., Russell, S., and Leike, J.,
  \enquote{Quantifying Differences in Reward Functions,} \emph{International
  Conference on Learning Representations}, 2021.

\bibitem[{Kendall(1938)}]{kendall1938new}
Kendall, M.~G., \enquote{A new measure of rank correlation,} \emph{Biometrika},
  Vol.~30, No. 1/2, 1938, pp. 81--93.

\end{thebibliography}

\end{document}